\documentclass{article}

\usepackage[final]{neurips_2025}

\usepackage[utf8]{inputenc} 
\usepackage[T1]{fontenc}    
\usepackage{hyperref}       
\usepackage{url}            
\usepackage{booktabs}       
\usepackage{amsfonts}       
\usepackage{nicefrac}       
\usepackage{microtype}      
\usepackage{xcolor}         
\usepackage{array}
\usepackage{multirow}
\usepackage{adjustbox}
\usepackage{amsmath} 
\usepackage{amsthm}
\usepackage[table]{xcolor} 

\usepackage{tcolorbox}
\definecolor{maroon}{cmyk}{0,0.87,0.68,0.32}
\definecolor{lightblue}{HTML}{DAE8FC}

\usepackage[normalem]{ulem}
\usepackage{fontawesome5}
\usepackage[ruled,vlined]{algorithm2e}
\usepackage{titletoc}
\usepackage{amssymb}
\usepackage{booktabs}
\usepackage{titlesec}

\titlecontents{section}
  [1.8em]                       
  {}                            
  {\bfseries\contentslabel{2.3em}} 
  {\bfseries}                   
  {\titlerule*[0.6pc]{.}\contentspage} 
  [\addvspace{12pt}]
\titlecontents{subsection}
  [3.8em]
  {}
  {\bfseries\contentslabel{3.2em}} 
  {\bfseries}
  {\titlerule*[0.6pc]{.}\contentspage}
  [\addvspace{6pt}]

\title{Unveiling the Spatial-temporal Effective Receptive Fields of Spiking Neural Networks}

\author{%
  Jieyuan Zhang$^{1}$, \ 
  Xiaolong Zhou$^{1}$, \
  Shuai Wang$^{1}$ \  
  Wenjie Wei$^{1}$, \
  Hanwen Liu$^{1}$, \\ [0.3em]
  \textbf{Qian Sun}$^{1}$, \ 
  \textbf{Malu Zhang}$^{1,3}$\thanks{Corresponding author: \href{mailto:maluzhang@uestc.edu.cn}{\faEnvelope[regular]maluzhang@uestc.edu.cn}},  
  \textbf{Yang Yang}$^{1}$,  
  \textbf{Haizhou Li}$^{2,3}$ \\ [0.3em]
  $^{1}$University of Electronic Science and Technology of China,\\ [0.3em]
  $^{2}$The Chinese University of Hong Kong, Shenzhen , \\ [0.3em]
  $^{3}$Shenzhen Loop Area Institute
}

\begin{document}

\maketitle

\begin{abstract}
    Spiking Neural Networks (SNNs) demonstrate significant potential for energy-efficient neuromorphic computing through an event-driven paradigm. While training methods and computational models have greatly advanced, SNNs struggle to achieve competitive performance in visual long-sequence modeling tasks. In artificial neural networks, the effective receptive field (ERF) serves as a valuable tool for analyzing feature extraction capabilities in visual long-sequence modeling. Inspired by this, we introduce the Spatio-Temporal Effective Receptive Field (ST-ERF) to analyze the ERF distributions across various Transformer-based SNNs. Based on the proposed ST-ERF, we reveal that these models suffer from establishing a robust global ST-ERF, thereby limiting their visual feature modeling capabilities. To overcome this issue, we propose two novel channel-mixer architectures: \underline{m}ulti-\underline{l}ayer-\underline{p}erceptron-based m\underline{ixer} (MLPixer) and \underline{s}plash-and-\underline{r}econstruct \underline{b}lock (SRB). These architectures enhance global spatial ERF through all timesteps in early network stages of Transformer-based SNNs, improving performance on challenging visual long-sequence modeling tasks. 
    Extensive experiments conducted on the Meta-SDT variants and across object detection and semantic segmentation tasks further validate the effectiveness of our proposed method.
    Beyond these specific applications, we believe the proposed ST-ERF framework can provide valuable insights for designing and optimizing SNN architectures across a broader range of tasks. The code is available at \href{https://github.com/EricZhang1412/Spatial-temporal-ERF}{\faGithub~EricZhang1412/Spatial-temporal-ERF}.
\end{abstract}

\section{Introduction}
    Spiking Neural Networks (SNNs)~\cite{MAASS19971659, gerstner2002spiking} have emerged as a prominent research focus, characterized by binary spike activation that offers high sparsity, event-driven processing~\cite{zhang2021rectified, zhang2025toward}, and biological plausibility~\cite{1257420}. Recent advances in encoding schemes~\cite{yu2014brain,guo2021neural,Qiu_Zhu_Chou_Wang_Deng_Li_2024}, training methodologies~\cite{wu2018spatio,zhang2025memory}, and neuromorphic hardware~\cite{davies2018loihi,pei2019towards,ma2024darwin3} have enabled SNNs to achieve remarkable success in diverse tasks, including image processing~\cite{10378556,lei2024spike2former,10.1007/978-3-031-73411-3_15}, point/event analysis~\cite{Qiu_Yao_Zhang_Chou_Qiao_Zhou_Xu_Li_2025,shan2025sdtrackbaselineeventbasedtracking}, language understanding~\cite{zhu2023spikegpt,xing2025spikellm, wang2025training}, and speech processing~\cite{zhang2024spikebased,wang2024global,wang2025ternary}. Nonetheless, SNNs still struggle to achieve performance comparable to their Artificial Neural Networks (ANNs) counterparts in visual long-sequence modeling tasks.

    \begin{figure}[htpb]
        \centering
        \includegraphics[width=1.0\linewidth]{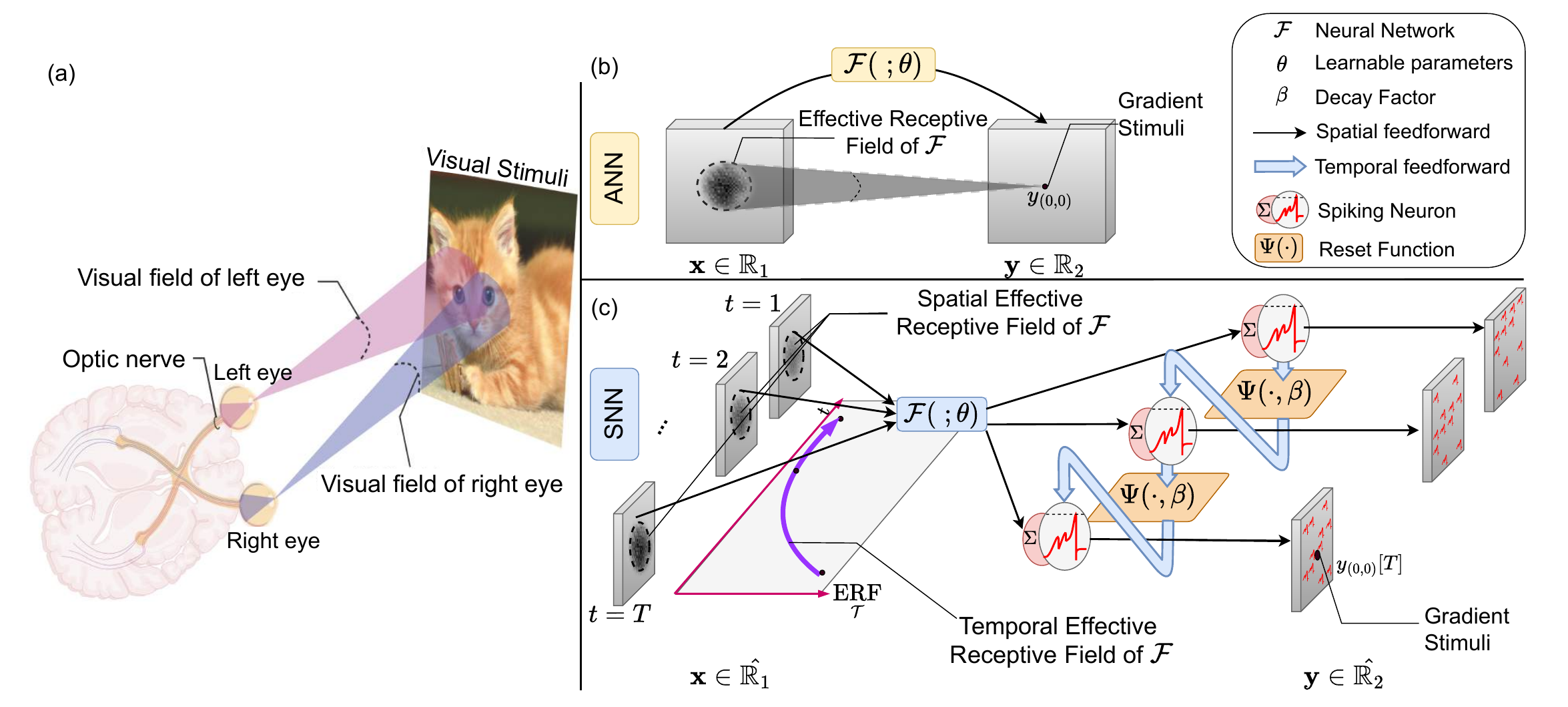}
        \caption{(a) The human visual field. (b) ERF in ANNs. (c) ST-ERF in SNNs. It extends the ERF to the temporal dimension, thus facilitating a comprehensive analysis of feature extraction in SNNs.}
        \label{fig:erf_vs_sterf}
    \end{figure}
    Compared to conventional image classification, visual long-sequence modeling tasks~\cite{yu2025mambaout,7298965} demand spatially dense outputs with prediction scales several orders of magnitude higher. This paradigm requires architectures capable of modeling long-range spatial dependencies, which are essential for achieving competitive performance~\cite{yu2025mambaout}. The Transformer~\cite{vaswani2017transformer} architecture introduces a self-attention mechanism that enables effective modeling of long-range spatial dependencies~\cite{dosovitskiy2020vit,SETR,10.1007/978-3-030-58452-8_13}.
    Motivated by this, recent studies have proposed various Transformer-based SNNs~\cite{zhou2023spikformer,zhou2023spikingformerspikedrivenresiduallearning,yao2025scaling,wang2025spiking}, achieving notable progress in visual long-sequence tasks~\cite{lei2024spike2former}. However, simply combining Transformers with SNNs may lead to suboptimal designs without fully considering the intrinsic spatio-temporal dynamics of spiking neurons. 
    To bridge this gap, a more structured and interpretable framework is required to examine how SNNs model spatio-temporal dependencies. In this context, receptive field (RF) analysis offers a concrete lens through which their feature extraction capacity and attention allocation can be theoretically characterized.
    
    In neuroscience, the RF represents the region of sensory input that can modulate a neuron's activity~\cite{hubel1962receptive}. 
    Borrowing this concept, the deep learning community defines a neuron's RF as the region of the input that can influence its output~\cite{le2017receptive}, with its size determined by network topology.
    However, this topology-based definition treats all positions within the RF equally, ignoring the learnable weights that determine the actual contribution of each input location. 
    To refine this problem, researchers proposed the effective receptive field (ERF)~\cite{NIPS2016_c8067ad1} to quantify input features' contributions to output features via gradient analysis.
    Unlike topology-based RF, gradient-based ERF provides a more faithful characterization of the network's feature extraction patterns. 
    However, such framework cannot be directly applied to SNNs due to the intrinsic spatio-temporal dynamics of spiking neurons. 
    Therefore, we introduce the Spatial-Temporal Effective Receptive Field (ST-ERF) framework, which quantifies input feature contributions across spatio-temporal locations to characterize SNNs' feature extraction patterns. By jointly modeling temporal dependencies and spatial relationships, ST-ERF facilitates a comprehensive analysis of information processing in SNNs.

    Based on the proposed ST-ERF, we analyze various Transformer-based SNNs and identify that existing models fail to establish effective global receptive fields across all timesteps.
    This limitation stems from the prevalent use of convolutional channel-mixers, which inherently introduce locality bias~\cite{10.5555/3666122.3669371}. 
    Despite facilitating efficient local feature extraction and long-range sparse modeling, this architectural design fundamentally constrains the long-range dense spatial interactions necessary for effective visual long-sequence modeling in SNNs.
    Building on these insights, we propose two novel channel-mixer architectures: \underline{m}ulti-\underline{l}ayer-\underline{p}erceptron-based m\underline{ixer} (MLPixer) and \underline{s}plash-and-\underline{r}econstruct \underline{b}lock (SRB). 
    These designs use pixel-wise MLPs to keep spatial features separate when mixing channels, which reduces locality bias and improves the global receptive field in early stages of Transformer-based SNNs.
    Extensive experiments demonstrate the effectiveness of our methods on visual long-sequence tasks. Specially, on COCO 2017 object detection and ADE20K semantic segmentation, our Meta-SDT-Base \cite{yao2025scaling} with SRB achieves 48.9\% $\mathrm{AP^{b}_{50}}$ and 43.7\% mIoU, respectively, while maintaining a smaller model size. These results surpass state-of-the-art Transformer-based SNNs, thereby validating our ST-ERF analysis and further advancing SNNs in visual long-sequence modeling. Our main contributions are listed as follows:
    \begin{itemize}
    \item We propose the ST-ERF framework, extending the traditional ERF concept to the temporal dimension with rigorous mathematical formalization. ST-ERF systematically quantifies how input features at different spatial and temporal locations contribute to output features. This provides a theoretical tool for understanding and optimizing feature extraction in SNNs. 
    \item We analyze various Transformer-based SNNs using the ST-ERF framework, revealing a critical limitation that existing models fail to establish a global ERF across all timesteps. To overcome this issue, we introduce two novel channel mixer designs: MLPixer and SRB, enabling Transformer-based SNNs to fully exploit their global modeling potential of long-range dependencies.
    \item 
    We conduct extensive experiments on visual long-sequence modeling tasks and demonstrate that our method achieves superior performance.
    For instance, our Meta-SDT-Base with SRB achieves 48.9\% $\mathrm{AP^{b}_{50}}$ on COCO 2017 detection and 43.7\% mIoU on ADE20K segmentation. It significantly outperforms existing state-of-the-art Transformer-based SNNs while using a smaller model size. These results strongly support the validity of our ST-ERF theoretical analysis and demonstrate the effectiveness of the proposed architectural designs. 
    \end{itemize}
\section{Related Works}

\subsection{Receptive Field in Neural Networks}
The human visual system perceives the external world through the visual fields of both eyes. As illustrated in Figure \ref{fig:erf_vs_sterf}(a), each eye covers a specific region of the visual space, and these regions partially overlap in the center to enable binocular vision. 
Neurons along the visual pathway respond selectively to stimuli within their RFs. 
Over the past decades, the RF theory has profoundly influenced our understanding of how the brain filters and integrates visual information across spatial locations~\cite{MA2024102582}. Inspired by the RF theory, deep neural networks adopt a similar principle by characterizing hierarchical ERFs that capture progressively abstract representations of input data. As shown in Figure~\ref{fig:erf_vs_sterf}(b), the ERF~\cite{NIPS2016_c8067ad1} formalizes this process by analyzing how spatial stimuli contribute to network activations.
ERF has motivated extensive research at different architectural levels, from understanding basic operators~\cite{Hayase2023UnderstandingMA} to designing higher-level modules and network structures such as Adaptive Receptive Fields~\cite{wei2017learning}, RF-Next~\cite{gao2022rf}, and AutoRF~\cite{dong2023autorf}. RF-based analysis has also driven advances in computational efficiency for lightweight architectures, influencing the development of CNN-based MobileNet variants~\cite{mobilenetv1, sandler2018mobilenetv2, howard2019searching} and MLP-based networks such as MLP-Mixer~\cite{tolstikhin2021mlp} and TSMixer~\cite{ekambaram2023tsmixer}. Building on these insights, this work extends the ERF concept to SNNs, offering a theoretical framework for analyzing and optimizing their spatio-temporal feature extraction processes.

\subsection{Visual Long-sequence Modeling in SNNs}
Visual long-sequence modeling refers to tasks that require multiple predictions per image, rather than a single-label classification~\cite{ranftl2021vision}. These tasks mainly include detection, segmentation, video understanding, and so on~\cite{yu2025mambaout}. As these tasks involve modeling complex spatial and temporal dependencies, they demand architectures capable of capturing long-range contextual information. Transformer has become the dominant paradigm for visual long-sequence modeling owing to its global self-attention mechanism and flexible scalability. However, such models still suffer from high computational costs, primarily due to the quadratic complexity of self-attention, dense prediction requirements, and high-resolution inputs~\cite{10613466}. Recently, leveraging the sparse spike-driven nature of SNNs has emerged as a promising direction to mitigate these computational costs. 
Spike-driven Transformer series~\cite{yao2024spikedriven,yao2025scaling,qiu2025quantized} adapt the standard Metaformer into an SNN framework for object detection and semantic segmentation, demonstrating the feasibility of SNNs in dense prediction tasks. Spike2Former~\cite{lei2024spike2former} integrates normalized integer leaky-and-integrated firing (NI-LIF) neurons and spike-driven deformable attention to achieve competitive performance on segmentation benchmarks while maintaining low energy consumption. Despite these advancements, SNNs still lag behind ANNs in visual long-sequence modeling. This underscores the need for deeper investigation into SNNs' spatio-temporal bottlenecks and architectural optimization.

\section{Theoretical Analysis of Spatio-temporal Effective Receptive Field}

In this section, we first introduce the concept of ERF in conventional ANNs.
Subsequently, we extend this conventional ERF into the temporal dimension to characterize the ST-ERF in SNNs. Finally, we introduce a loss-derived method to efficiently compute ST-ERF in SNNs.

\subsection{ERF in ANNs}
The concept of the ERF has been widely adopted to analyze how input features contribute to network activations and how such influences are distributed within the RF~\cite{Hayase2023UnderstandingMA,wei2017learning}. Under the assumption of a single channel per layer, Luo et al.~\cite{NIPS2016_c8067ad1} mathematically characterized how each input feature contributes to the output of a neural network layer. It can be defined as follows:
\begin{equation}
    \mathrm{ERF}_{(i,j)}[y_{(m,n)};\mathbf{x}] = \frac{\partial y_{(m,n)}}{\partial x_{(i,j)}},
    \label{ERF_def}
\end{equation}
where $\mathbf{x} \in \mathbb{R}_{1}$ is the input feature and $\mathbf{y} \in \mathbb{R}_{2}$ is the output feature. In this manner, the ERF measures the partial derivative of an output feature $y_{(m,n)} \in \mathbf{y}$ with respect to each input feature $x_{(i,j)} \in \mathbf{x}$ within a given layer. As illustrated in Figure~\ref{fig:erf_vs_sterf}(b), the ERF of a given network \( \mathcal{F}(;\theta)\) describes the input regions that contribute to a particular output activation. 

As shown in Eq.~\eqref{ERF_def}, the ERF can be computed at any output location. However, most studies evaluate the ERF at the central output feature $y_{(0,0)}$ by assigning a unit gradient to this location~\cite{ding2022scaling, wang2025uniconvnet}. This practice establishes a centered and symmetric reference, ensuring stable and comparable visualization results. In this work, we also follow the setting of~\cite{NIPS2016_c8067ad1} and adopt the ERF at the central output feature $y_{(0,0)}$ as the evaluation metric.

\subsection{ST-ERF in SNNs}

Due to the inherent temporal dynamics, SNNs require additional consideration of the input at each timestep. To address this, we formally define the ST-ERF (i.e., $\mathrm{ERF}^{(\mathcal{S},\mathcal{T})}$). Firstly, we redefine the mapping relationship of SNNs. Consider a SNN layer with learnable parameters $\theta$ that maps input spike features $\mathbf{x}[1:T] \in \hat{\mathbb{R}_{1}}$ to output spike features $\mathbf{y}[1:T] \in \hat{\mathbb{R}_{2}}$:
\begin{equation} 
    \mathbf{y}[1:T] = \mathcal{F}(\mathbf{x}[1:T]; \theta), \mathcal{F}: \hat{\mathbb{R}_{1}} \to \hat{\mathbb{R}_{2}}.
\end{equation}
Its ERF needs to account not only for the accumulation across spatial dimensions but also for that across temporal dimensions. Specifically, $\mathrm{ERF}^{(\mathcal{S},\mathcal{T})} \in \hat{\mathbb{R}_1}$ can be expressed as:
\begin{equation}
    \mathrm{ERF}^{(\mathcal{S},\mathcal{T})}_{(i,j)}[\,y_{(m,n)}[t], \tau; \mathbf{x}\,]
    =
    \frac{\partial y_{(m,n)}[t]}{\partial x_{(i,j)}[t-\tau]}, 1 \leq t \leq T, 0 \leq \tau \leq t-1.
    \label{ST-ERF_def}
\end{equation}
Accordingly, $\mathrm{ERF}^{(\mathcal{S},\mathcal{T})}$ quantifies how much each input feature $x_{(i,j)}[t-\tau] \in \mathbf{x}$ at a previous timestep $t-\tau$ contributes to a specific output feature $y_{(m,n)}[t] \in \mathbf{y}$.
Based on this definition, the spatial ERF (i.e., $\mathrm{ERF}^{(\mathcal{S})}$) can be seen as the weighted average of the ST-ERFs over all timesteps:
\begin{equation}
    \mathrm{ERF}^{(\mathcal{S})}_{(i,j)}[\,y_{(m,n)}; \mathbf{x}] = \frac{1}{T}\sum_{t=1}^{T} \sum_{\tau=0}^{t-1} w(t,\tau) \cdot \mathrm{ERF}^{(\mathcal{S},\mathcal{T})}_{(i,j)}[\,y_{(m,n)}[t]; \mathbf{x}, \tau\,],
    \label{Spatial_sterf_def}
\end{equation}
where $w(t,\tau)$ represents the relative contribution of the input with delay $\tau$ at time $t$ to the output. The specific form of $w(t,\tau)$ depends on the neuronal dynamics and network architecture. For example, in Leaky Integrate-and-Fire (LIF) neurons, inputs closer to the current time step may have a higher influence due to the decay of membrane potential over time. 

The temporal ERF (i.e., $\mathrm{ERF}^{(\mathcal{T})}$) can be seen as the integration over the spatial dimensions of ST-ERF to indicate the contribution of inputs at different timesteps to the final output:
\begin{equation}
    \mathrm{ERF}^{(\mathcal{T})}[\tau; \mathbf{x}] = \sum_{i,j} \sum_{m,n} \mathrm{ERF}^{(\mathcal{S},\mathcal{T})}_{(i,j)}[\,y_{(m,n)}[T]; \mathbf{x}, \tau\,].
    \label{Temporal_STERF_def}
\end{equation}
As shown in Figure~\ref{fig:erf_vs_sterf}(c), we visualize an example of the ST-ERF. Similar to conventional ERF analysis, we focus on the center of the feature map at a specific timestep (e.g., the final timestep in Fig.~\ref{fig:erf_vs_sterf}(c)) to analyze the spatio-temporal feature representations in an SNN. Depending on the purpose of analysis, one may investigate the spatial distribution of the ST-ERF at a given timestep (spatial ERF) or its temporal distribution across one or more layers (temporal ERF). 

\subsection{Loss-Derived Calculation for ST-ERFs}
\label{LDC}
Based on Eq.~\eqref{ST-ERF_def}, computing the ST-ERF in SNNs requires evaluating first-order derivatives of outputs with respect to all input features. To obtain the ST-ERF conveniently, we introduce the loss-derived calculation method to efficiently compute using PyTorch's Automatic Differentiation functionality. 
Consider a SNN with input spike features $s^{\ell-1}$, output spike features $s^{\ell}$ at the $\ell$-th layer, and an arbitrary loss function $\mathcal{L}$. The spatial ERF of SNNs can be easily obtained by calculating the average of the gradient of the loss with respect to input features at position $(i,j)$ across all timesteps $T$. Specifically, it can be computed as follows: 
\begin{equation}
        \mathrm{ERF}^{(\mathcal{S})}_{(i,j)}[\,s^{\ell}_{(0,0)}]
         = \frac{1}{T}\sum_{t=1}^{T} \frac{\partial s^{\ell}_{(0,0)}[t]}{\partial s^{\ell-1}_{(i,j)}[t]} = \frac{1}{T} \sum_{t=1}^{T} \frac{\partial \mathcal{L}}{\partial s^{\ell-1}_{(i,j)}[t]}, 
         \text{when} \: \forall t, \frac{\partial \mathcal{L}}{\partial s^{\ell}_{(\hat{i},\hat{j})}[t]} =  
    \begin{cases}
        1, & \hat{i}=0,\hat{j}=0, \\
        0, & \text{otherwise}
    \end{cases}.
        \label{LD_SERF_Analysis}
\end{equation}
The temporal ERF of SNNs can be obtained by calculating the sum of the gradient of the loss function with respect to input features at timestep $T-\tau$ across all spatial positions. It can be computed as:
\begin{equation}
    \mathrm{ERF}^{(\mathcal{T})}[\tau] 
     = \sum_{i,j} \sum_{\hat{i},\hat{j}} \frac{\partial s_{(\hat{i},\hat{j})}^{\ell}[T]}{\partial s_{(i,j)}^{\ell-1}[T-\tau]} = \sum_{i,j} \frac{\partial \mathcal{L}}{\partial s_{(i,j)}^{\ell-1}[T-\tau]}, \quad
\text{when } \: \forall \hat{i},\hat{j}, \frac{\partial \mathcal{L}}{\partial \mathbf{s}^{\ell}_{(\hat{i},\hat{j})}[T]} = 1. 
    \label{LD_TERF_Analysis}
\end{equation}
Proof can be found in Appendix~\ref{proof_property_1}.  
We refer to the conditions in Eq.~\eqref{LD_SERF_Analysis} and~\eqref{LD_TERF_Analysis} as gradient stimuli.
Based on this proposition, we could easily obtain the spatial and temporal ERF with automatic back-propagation without an explicit loss function.

\section{Problem Analysis on Transformer-based SNNs using ST-ERF}
\label{section 4}
In this section, we use the ST-ERF framework to analyze existing Transformer-based SNNs and identify their limitations in visual long-sequence modeling tasks.
\subsection{Different ST-ERF Behaviors in Transformer-based SNNs}

We apply the ST-ERF framework to analyze Transformer-based SNNs' spatial ERF behaviors across all timesteps. Specifically, we compared two groups of architectures(a: ViT-like architecture group and b: Meta-architecture group) with their ANN counterparts to investigate the differences in the formation of their spatial ERFs. For the loss-derived calculation, we set the central patch across all channels and timesteps in the output tensor as the gradient stimuli (uniform values of 1), then perform automatic back-propagation. Each experiment comprised 60 iterations using randomly sampled input tensors under standard normal distribution ($\mu = 0, \sigma^2 = 1$). Note that we average the ST-ERF over all timesteps to obtain a clear visualization. 
\begin{figure}[!ht]
    \centering
    \includegraphics[width=1.0\linewidth]{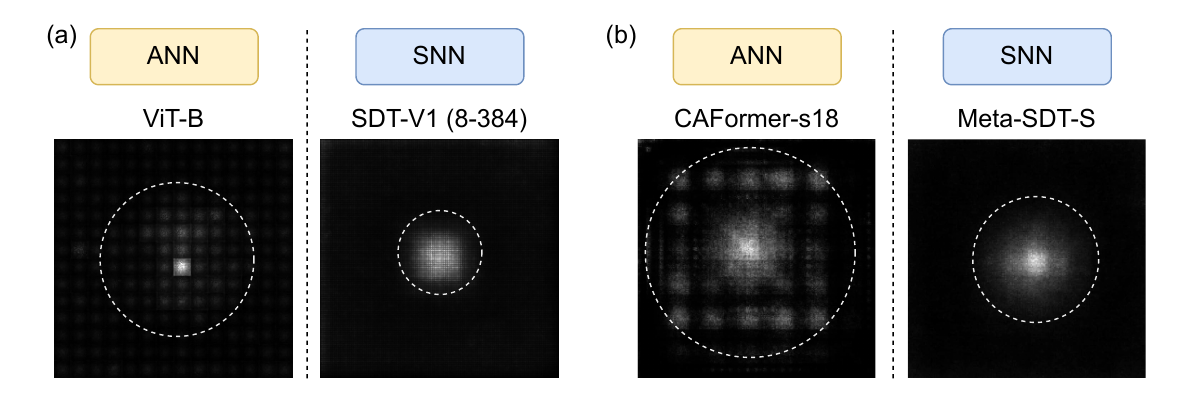}
    \caption{Comparison of spatial ERF with ANN Vision Transformers and ST-ERF with different Transformer-based SNNs. (a) ViT-like architecture comparison group:  ViT-B and SDT-V1. (b) Meta-architecture comparison group: CAFormer-s18 and its counterpart Meta-SDT-S.}
    \label{fig:serf_viscomp}
    \vspace{-10pt}
\end{figure}

The comparison of ViT-like architectures is illustrated in Figure~\ref{fig:serf_viscomp}(a). Compared with the classic ViT-B, SDT-V1 exhibits a more centrally concentrated yet markedly narrower spatial ERF. This observation suggests that the architectural modifications in SDT-V1 may restrict the receptive field’s spatial extent, thereby enhancing its attention on local spatial dependencies. 
In fact, SDT-V1 adopts a fundamentally different strategy from the vanilla ViT in the patch splitting stage. Specifically, SDT-V1 employs the Spike Patch Splitting (SPS) module, consisting a Patch Splitting Module (PSM) to linearly project the input image and a Relative Position Embedding (RPE)~\cite{yao2023spikedriven} block to generate the latent position information. The SPS module incorporates multiple convolutional layers at the early stage of the network, facilitating low-level spatial features extraction from input images.

The comparison of meta-architectures is illustrated in Figure~\ref{fig:serf_viscomp}(b). Although Meta-SDT exhibits ERF behaviors similar to those of its ANN counterparts, it also struggles to maintain long-range feature attention. This limitation can be attributed to the additional employment of convolutional layers, which tend to emphasize localized features rather than global spatial contexts. Compared with CAFormer, Meta-SDT introduces the Re-parameterization Convolution (RepConv)\cite{ding2021repvgg} to perform the linear projection of queries, keys, and values\cite{yao2024spikedriven}. This design enhances local feature extraction, yet it inherently constrains the model’s capacity to aggregate information across distant spatial regions. Together, these findings suggest that the convolutional operations enhances local feature sensitivity but poses challenges for maintaining long-range spatial coherence in Transformer-based SNNs.

\subsection{Visual Long-sequence Modeling Needs Global ST-ERF}
Visual long-sequence modeling tasks often involve dense predictions across an entire image, requiring the processing of thousands of input tokens~\cite{yu2025mambaout}.
Therefore, capturing long-range dependencies and global context is crucial for achieving accurate and robust representations~\cite{NEURIPS2024_baa2da9a}. 
Prior studies have found that vision models with global receptive fields often excel at segmentation and detection, for instance when using self-attention mechanisms as in Transformer architectures~\cite{NEURIPS2024_baa2da9a,Qi2025}.
In contrast, architectures lacking global context integration tend to struggle. While early convolutional layers excel at extracting low-level structural patterns~\cite{li2023middle}, their locality inherently limits the capacity to capture long-range dependencies, making them suboptimal for visual long-sequence tasks. 

However, despite the need for global spatial awareness in visual long-sequence modeling, Transformers-based SNNs still fail to achieve a truly global ST-ERF. As discussed above, they tend to focus heavily on the center and expand to limited size. 
This contrasts sharply with the expected behavior required for visual long-sequence modeling tasks, where the weak global ST-ERF limits information aggregation and consequently degrades performance on such scenarios~\cite{NEURIPS2024_baa2da9a}.

\section{Methods}
In this section, we propose two novel channel-mixing designs, MLPixer and SRB, which enable Transformer-based SNNs to more effectively capture long-range dependencies. Furthermore, we integrate these modules into the Meta-SDT architecture to enhance performance on visual long-sequence modeling tasks.

\subsection{Design of Channel Mixer Block}
\begin{figure}[!ht]
    \centering
    \includegraphics[width=1.0\linewidth]{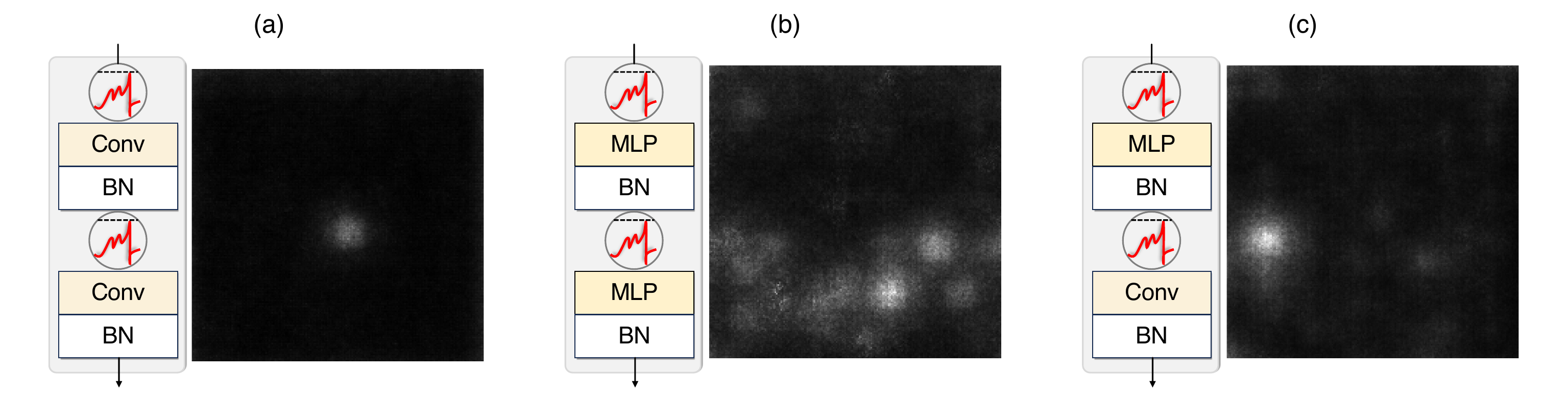}
    \caption{Comparison between the original channel mixer design and our proposed methods, along with their ST-ERF. For clearer visualization, the ST-ERF is averaged over timesteps. (a) Vanilla convolution-based channel mixer. (b) Proposed MLPixer architecture. (c) Proposed SRB architecture. Obviously, the vanilla convolution-based channel mixer exhibits a limited ST-ERF, whereas our MLPixer and SRB modules achieve a more global ST-ERF. Moreover, due to the reduced use of convolutions, MLPixer exhibits an even broader effective receptive field.}
    \label{fig:meth_cmcomp}
    \vspace{-10pt}
\end{figure}

To enhance the global modeling capability of SNN in visual long-sequence tasks, we propose two novel channel mixer designs. The first is the multi-layer perceptron-based mixer (MLPixer), which employs a two-layer MLP structure to more effectively extract global features. It is defined as follows:
\begin{equation}
\mathrm{MLPixer}(\mathbf{X}) 
= \mathrm{BN} \Big(
    \mathrm{MLP} \big(
        \mathbb{SN} \{
            \mathrm{BN}(
                \mathrm{MLP}\{
                    \mathbb{SN}(\mathbf{X})
                \}
            )
        \}
    \big)
\Big),
\end{equation}
where \(\textbf{X} \in \mathbb{R}^{T\times B \times N \times D}\) denotes the input of channel mixers in the Transformer block. $\mathbb{SN}(\cdot)$ denotes a spiking neuron layer that transforms the input sequence into the spike trains. \(\mathrm{MLP}(\cdot)\) denotes a single-layer fully connected (FC) operation, and \(\mathrm{BN}(\cdot)\) denotes batch normalization. 

Compared with vanilla channel mixers~\cite{yu2024metaformer, yao2024spikedriven} that rely on convolution operations, the MLPixer employs a two-layer MLP operation to mix features across channels. This design reduces reliance on convolutional operations, mitigating the ERF's bias toward a Gaussian-like central concentration and enabling SNNs to capture long-range dependencies more effectively.

Building on this, we further propose the SRB module. It replaces only the second convolution in the channel mixer with a single-layer MLP operation. Specifically, the SRB is defined as follows:
\begin{equation}
\mathrm{SRB}(\mathbf{X}) 
= \mathrm{BN} \Big(
    \mathrm{MLP} \big(
        \mathbb{SN} \{
            \mathrm{BN}(
                \mathrm{Conv}\{
                    \mathbb{SN}(\mathbf{X})
                \}
            )
        \}
    \big)
\Big).
\end{equation}
Here, \(\text{Conv}(\cdot)\) denotes a 1$\times$1 convolution operation. In this manner, SRB module reduces additional parameters while maintaining performance. To validate the effectiveness of our approach, we visualize the ERFs of the Conv-based mixer, the MLPixer, and the SRB modules. 

As shown in Figure~\ref{fig:meth_cmcomp}(a), the vanilla convolution-based channel mixer exhibits a limited ST-ERF. In contrast, the proposed MLPixer and SRB modules demonstrate a more global ST-ERF. Furthermore, the comparison between Figure~\ref{fig:meth_cmcomp}(b) and Figure~\ref{fig:meth_cmcomp}(c) further demonstrates that MLPixer exhibits a more global ERF. This stems from reduced use of convolutions and further suggests that MLPs provide stronger global modeling capacity than convolutional operators. We will validate the proposed module on visual long-sequence modeling tasks in the experiment section.

\begin{figure}[!ht]
    \centering
    \includegraphics[width=1.0\linewidth]{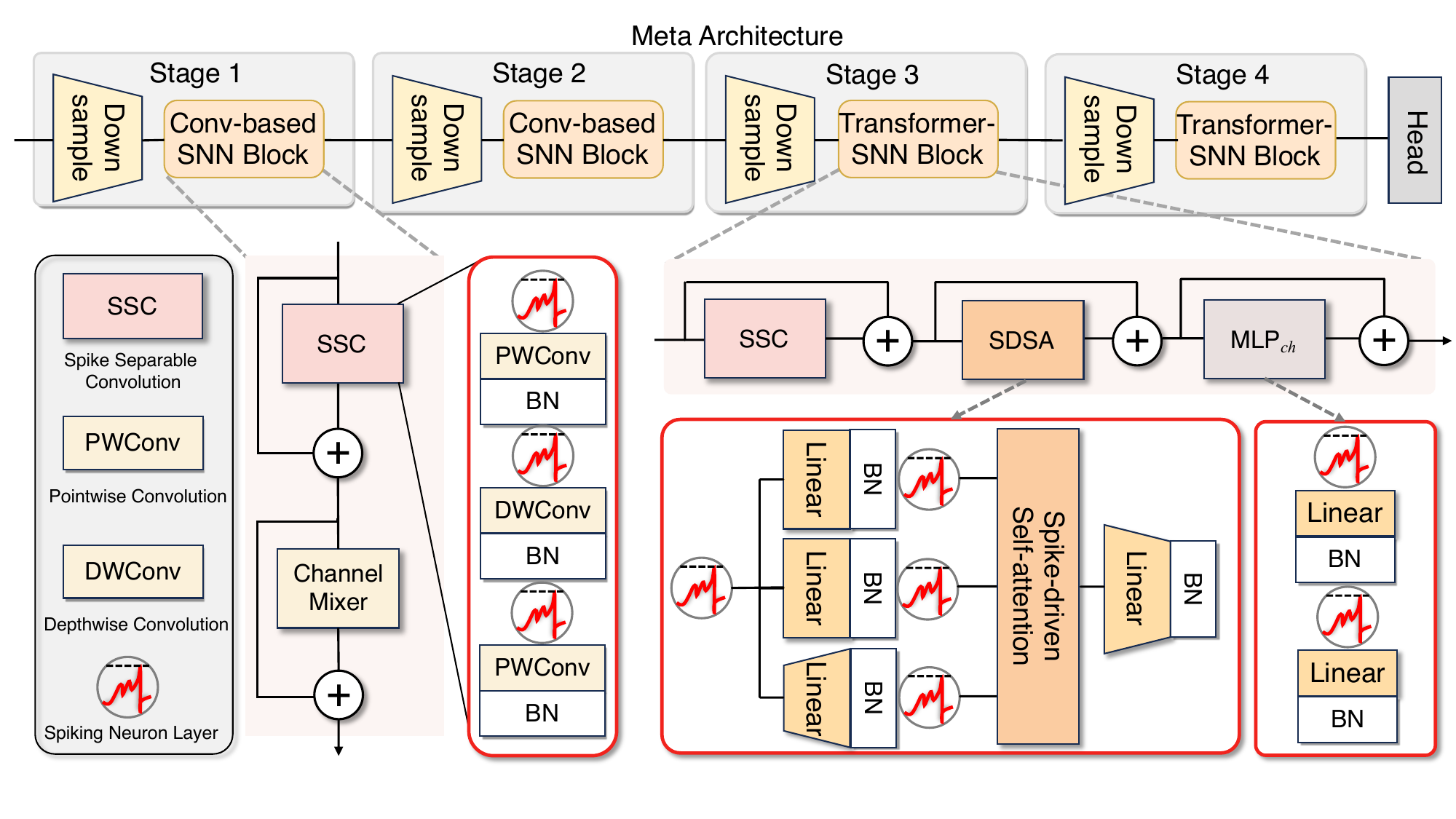}
    \caption{The overall architecture of Meta-SDT, which typically comprises four hierarchical stages. The first two stages use convolution-based SNN blocks, while the latter two adopt Transformer-SNN blocks. To strengthen the global modeling capacity of SNNs, we introduce two novel channel mixer architectures, MLPixer and SRB, to replace the convolution-based SNN blocks in the first two stages.}
    \label{fig:meth_viscomp}
\end{figure}

\subsection{Overall Architecture}
To further validate the effectiveness of our approach, we integrate the proposed SRB and MLPixer modules into the CAFormer~\cite{yu2024metaformer} and Meta-SDT~\cite{yao2025scaling} architectures. As shown in Figure~\ref{fig:meth_viscomp}, these architectures adopt a multi-stage design, where the first two stages consist of Conv-based SNN blocks, and the latter two stages comprise Transformer-SNN blocks. In this work, we replace only the operations in the first two stages. Specifically, the first two stages are represented as:
\begin{equation}
\label{eq:conv-mlp-snn-block}
\begin{aligned}
    &\mathbf{X}' = \mathbf{X} + \mathrm{SSC}(\mathbf{X}), \mathbf{X}''    = \mathbf{X}'  + \mathrm{Mixer}_{\epsilon}(\mathbf{X}').
\end{aligned}
\end{equation}
Here, $\mathrm{Mixer}_{\epsilon}(\cdot)$ is the channel mixer. In this work, we implement the approach using both the MLPixer module and the SRB module.  Similar to the vanilla channel mixer, our method adopts an up-projection followed by a down-projection with a nonlinear activation in between, where $\epsilon > 1$ represents the intermediate dimensional expansion ratio. $\mathrm{SSC}(\cdot)$ is the spike-driven separable convolution block as token mixer, it is defined as follows:
\begin{equation}
\mathrm{SSC}(\mathbf{X})
= \mathrm{PWConv}_2\!\left(
    \mathbb{SN}\!\left(
        \mathrm{DWConv}\!\left(
            \mathbb{SN}\!\left(
                \mathrm{PWConv}_1\!\left(
                    \mathbb{SN}(\mathbf{X})
                \right)
            \right)
        \right)
    \right)
\right).
\end{equation}
$\mathrm{PWConv}_1(\cdot)$ and $\mathrm{PWConv}_2(\cdot)$ are pointwise convolutions, $\mathrm{DWConv}(\cdot)$ is depthwise convolution. $\mathbb{SN}(\cdot)$ denotes the spiking neuron layer. To maintain the spike-driven characteristics of the network, we implement membrane-shortcut residual connection mechanism. Furthermore, Transformer-SNN blocks are utilized in Stage 3 and Stage 4, following the same configuration as that of Meta-SDT-V3~\cite{yao2025scaling}. We will further verify the effectiveness of the proposed method in the experimental section.

\section{Experiments}
In this section, we validate the effectiveness of our method through visualization and experimental analysis. First, we examine the changes in the ST-ERF after integrating the proposed modules into Meta-SDT, showing that our method achieves stronger global spatial receptive fields across all stages. Second, we evaluate its performance improvement on long-sequence modeling tasks, including object detection and semantic segmentation. Finally, we further investigate the method on complex event modeling tasks to assess its applicability in more challenging scenarios.

\subsection{ST-ERF Behavior in Transformer-based SNNs}

In order to study the impact of our proposed block on the receptive field of Meta-SDT, we compared temporal-averaged spatial ERFs between our two Meta-SDT variants with previous models. 
We initialized the central spatial feature across all channels and timesteps in the output tensor as uniform gradient stimuli (value = 1), and propagated the gradients backward through the network. Each experiment consisted of 60 iterations with input tensors randomly drawn from a standard normal distribution ($\mu = 0, \sigma^2 = 1$).

\begin{figure}[!ht]
    \centering
    \includegraphics[width=1.0\textwidth]{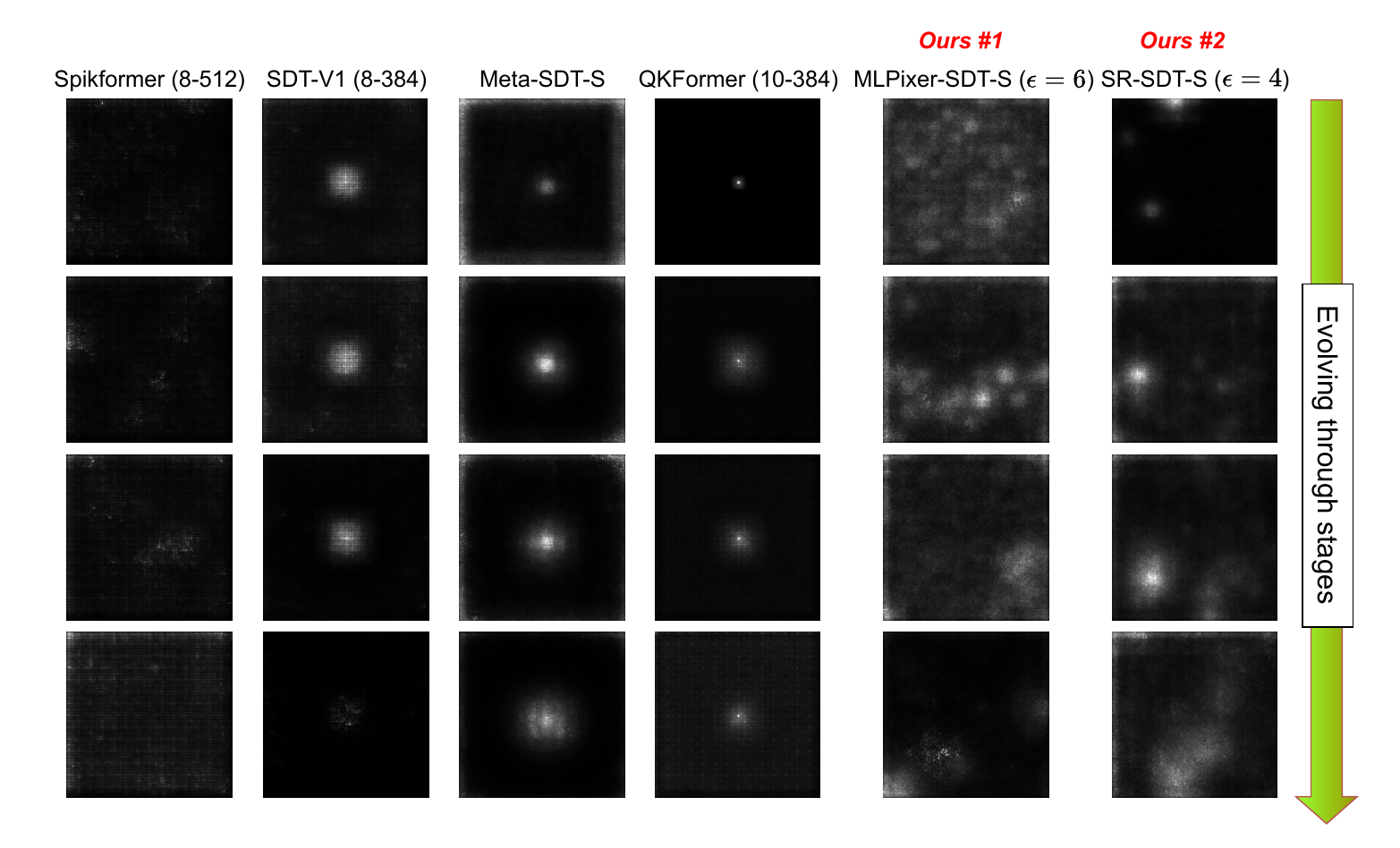}
    \caption{Comparison of temporal-averaged spatial ERF evolution across stages. From top to bottom are Stage 1 through Stage 4. Spikformer shows diffuse receptive fields across all stages. SDT-V1, Meta-SDT, and QKFormer exhibit more centered spatial distributions that gradually expand as depth increases. Our two Meta-SDT variants establish global spatial receptive fields in the early stages.
    }
    \vspace{-0.5cm}
    \label{fig:comp_different_svit_erf_s}
\end{figure}

The results are illustrated in Figure \ref{fig:comp_different_svit_erf_s}. Surprisingly, we found that Spikformer exhibits diffuse receptive fields across all stages. The SDT-V1, Meta-SDT, and QKFormer demonstrate markedly centered distribution that gradually expand as the network deepens, all manifesting a Gaussian-like effect. Additionally, we observed  dissipation of spatial ERF in SDT-V1 during the final stage. In contrast, our proposed two Meta-SDT variants establish robust global spatial receptive fields in the early stages. The MLPixer-SDT establishes a strong global spatial ERF in Stage 1. As the network deepens, its spatial ERF selectively contracts toward specific regions. The ERF behavior in SRB-SDT is slightly different, as it only begins to form a preliminary spatial ERF at Stage 2, and this distribution continues to evolve with increasing network depth.

\subsection{Performance in Visual Long-sequence Modeling Tasks}
We selected two challenging datasets to evaluate performance on classic visual long-sequence modeling tasks: object detection and instance segmentation on COCO 2017, and semantic segmentation on the ADE20K dataset. We choose the Meta-SDT(v3)~\cite{yao2025scaling} as the baseline and construct Meta-SDT variants with MLPixer($\epsilon4$), MLPixer($\epsilon6$) and SRB($\epsilon4$). 

\textbf{Performance on COCO 2017} \space We evaluate the efficacy of the MLPixer and SRB on Meta-SDT and select the classic and large-scale COCO~\cite{lin2014ms-coco} dataset as our benchmark for evaluation. Following the previous work~\cite{yao2024spikedriven,yao2025scaling,qiu2025quantized}, we use the MMdetection~\cite{chen1906mmdetection} codebase with a spiking version and then deploy our model. We employ the Meta-SDT~\cite{yao2025scaling} with two variants as the backbone network to extract features, along with fine-tuning Mask R-CNN~\cite{he2017mask} for object detection and instance segmentation. All backbone networks are pretrained on ImageNet-1K~\cite{deng2009imagenet}, while the incremental layers are initialized following~\cite{glorot2010understanding}. During fine-tuning, we strictly obey the $1 \times$ training schedule. 

The comparison results of object detection and instance segmentation are shown in Table~\ref{table_detection}. Under the same training schedule, both the MLPixer and SRB variants outperforms the baseline across all metrics. More specifically, the SRB variant exceeds the performance of SDTv3-T and SDTv3-B by $10.42\%$ and $4.26\%$ on the $ \mathrm{ AP^{b}_{50} }$ metric, while maintaining almost the same model size. In conclusion, our approach demonstrates efficacy in object detection and instance segmentation, setting a new benchmark for COCO dataset in the SNN domain.

\textbf{Performance on ADE20K} \space We evaluate the performance of MLPixer and SRB on the semantic segmentation task using the challenging ADE20K dataset \cite{zhou2019ade20k}. Similar to the COCO experiments, we utilize the spiking version of MMSegmentation \cite{contributors2020mmsegmentation} as our codebase and employ the Meta-SDT~\cite{yao2025scaling} with two variants as the backbone network. We fine-tune the Semantic FPN framework \cite{kirillov2019semetic-fpn} for semantic segmentation. Backbone networks are initialized with ImageNet-1K pre-trained weights \cite{deng2009imagenet}, and new layers follow the initialization scheme of \cite{glorot2010understanding}. All models are strictly obey the same training schedule for 160k iterations. 

As shown in Table~\ref{tab:finetune-ADE20K}, both MLPixer and SRB variants surpass the baseline in terms of mIoU. The SRB variant improves performance by 3.3\% and 2.6\% over SDTv3-T and SDTv3-B, respectively, while reducing parameters by 0.3M and 1.2M. The MLPixer($\epsilon4$) variant achieves the largest parameter reduction of 0.6M and 2.4M, with comparable or superior accuracy to SDTv3-T and SDTv3-B. These results highlight the effectiveness of the proposed modules in enhancing semantic segmentation on ADE20K.

\begin{table}[!ht]
\centering
\begin{minipage}[t]{0.53\textwidth}
\centering
\footnotesize 
\setlength{\tabcolsep}{1pt} 
\begin{tabular}{l cc cccccc}
\toprule
Arch. & \#T & \#P & $ \mathrm{ AP^{b} }$ & $ \mathrm{ AP^{b}_{50} }$ & $ \mathrm{ AP^{b}_{75} }$ & $ \mathrm{ AP^{m} }$ & $ \mathrm{ AP^{m}_{50} }$ & $ \mathrm{ AP^{m}_{75} }$ \\ 

\midrule
SDTv3-T\cite{yao2025scaling} & 4 & 25M & 15.2 & 35.5 & 10.2 & 15.2 & 33.0 & 12.3 \\
\rowcolor{lightblue!50}
MLPixer\scriptsize ($\epsilon4$) & 4 & 24M & 16.2 & 37.0 & 11.5 & 15.2 & 32.9 & 12.5 \\
\rowcolor{lightblue!50}
MLPixer\scriptsize ($\epsilon6$) & 4 & 25M & 17.5 & 38.5 & 13.2 & 16.2 & 34.5 & 13.5 \\
\rowcolor[HTML]{DAE8FC}
SRB\scriptsize ($\epsilon4$) & 4 & 25M & \textbf{18.2} & \textbf{39.2} & \textbf{13.8} & \textbf{17.5} & \textbf{34.8} & \textbf{14.3} \\
\midrule
SDTv3-B\cite{yao2025scaling} & 4 & 39M & 21.7 & 46.9 & 17.0 & 20.1 & 41.8 & 17.5 \\
\rowcolor{lightblue!50}
MLPixer\scriptsize ($\epsilon4$) & 4 & 36M & 22.9 & 47.6 & 19.2 & 21.0 & 43.4 & 18.3 \\
\rowcolor{lightblue!50}
MLPixer\scriptsize ($\epsilon6$) & 4 & 39M & 25.1 & 48.8 & 22.5 & 21.9 & 43.5 & 19.6 \\
\rowcolor[HTML]{DAE8FC}
SRB\scriptsize ($\epsilon4$) & 4 & 37M & \textbf{25.8} & \textbf{48.9} & \textbf{22.8} & \textbf{22.5} & \textbf{43.9} & \textbf{20.4} \\
\midrule
\end{tabular}
\caption{Object detection and instance segmentation with Mask R-CNN on COCO val2017, using ImageNet-1K pre-training and $1 \times$ training schedule. }
\label{table_detection}
\end{minipage}
\hfill
\begin{minipage}[t]{0.44\textwidth}
\centering
\footnotesize
\setlength{\tabcolsep}{2pt}
\begin{tabular}{lcccc}
\toprule
    Arch. & Ch. Mixer & \#T & Param.(M) &mIoU(\%)\\
   \midrule
   \multirow{4}{*}{\parbox{0.9cm}{SDTv3\\-T\cite{yao2025scaling}}} 
   & C2d-k3\scriptsize ($\epsilon4$) & 4 & 6.5~\textbf{\textcolor{black}{\scriptsize BASE}} & 34.9~\textbf{\textcolor{black}{\scriptsize BASE}}\\
   & \cellcolor{lightblue!50}MLPix.\scriptsize ($\epsilon4$) & \cellcolor{lightblue!50}4 & \cellcolor{lightblue!50}5.9~\textbf{\textcolor{blue}{\scriptsize ($\downarrow$0.6)}} & \cellcolor{lightblue!50}34.9~\textbf{\textcolor{red}{\scriptsize ($\uparrow$0.0)}}\\
   & \cellcolor{lightblue!50}MLPix.\scriptsize ($\epsilon6$) & \cellcolor{lightblue!50}4 &\cellcolor{lightblue!50}6.6~\textbf{\textcolor{red}{\scriptsize ($\uparrow$0.1)}} & \cellcolor{lightblue!50}35.9~\textbf{\textcolor{red}{\scriptsize ($\uparrow$1.0)}}\\
   & \cellcolor{lightblue}SRB\scriptsize ($\epsilon4$) & \cellcolor{lightblue}4  & \cellcolor{lightblue}6.2~\textbf{\textcolor{blue}{\scriptsize ($\downarrow$0.3)}} & \cellcolor{lightblue}\textbf{38.2}~\textbf{\textcolor{red}{\scriptsize ($\uparrow$3.3)}}\\
   \midrule

   \multirow{4}{*}{\parbox{0.9cm}{SDTv3\\-B\cite{yao2025scaling}}} 
   & C2d-k3\scriptsize ($\epsilon4$) & 4 & 20.4~\textbf{\textcolor{black}{\scriptsize BASE}} & 41.1 ~\textbf{\textcolor{black}{\scriptsize BASE}}\\
   & \cellcolor{lightblue!50}MLPix.\scriptsize ($\epsilon4$) & \cellcolor{lightblue!50}4 & \cellcolor{lightblue!50}18.0~\textbf{\textcolor{blue}{\scriptsize ($\downarrow$2.4)}} & \cellcolor{lightblue!50}42.0~\textbf{\textcolor{red}{\scriptsize ($\uparrow$0.9)}}\\
   & \cellcolor{lightblue!50}MLPix.\scriptsize ($\epsilon6$) & \cellcolor{lightblue!50}4 & \cellcolor{lightblue!50}20.7~\textbf{\textcolor{red}{\scriptsize ($\uparrow$0.3)}} & \cellcolor{lightblue!50}43.4~\textbf{\textcolor{red}{\scriptsize ($\uparrow$2.3)}}\\
   & \cellcolor{lightblue}SRB\scriptsize ($\epsilon4$) & \cellcolor{lightblue}4 & \cellcolor{lightblue}19.2~\textbf{\textcolor{blue}{\scriptsize ($\downarrow$1.2)}} & \cellcolor{lightblue}\textbf{43.7}~\textbf{\textcolor{red}{\scriptsize ($\uparrow$2.6)}}\\
   \midrule
\end{tabular}
\caption{Segmentation results on ADE20K based on different mixer block, using ImageNet-1K pre-training and 160k iter.}
\label{tab:finetune-ADE20K}
\end{minipage}
\end{table}

\subsection{Performance in Complex Event Modeling Tasks}

\textbf{Event-based Tracking} \space
We evaluate the effectiveness of two channel mixers in the context of event-based tracking, a highly challenging yet practically significant application domain for SNNs. Our experiments follow the SDTrack pipeline~\cite{shan2025sdtrackbaselineeventbasedtracking}, which employs the Global Trajectory Prompt method to convert event streams into event frames. We strictly adhere to the original training protocol, modifying only the backbone by replacing SDTrack with our proposed SDTrack+MLPixer or SDTrack+SRB variants. As presented in Table~\ref{tab:fe108_visevent_tracking}, extensive experiments on the FE108~\cite{zhang2021object} and VisEvent~\cite{wang2021viseventbenchmark} datasets demonstrate that our architectures surpass the original SDTrack in several key metrics. These results confirm that both the MLPixer and SRB designs preserve the Transformers-based SNNs' performance, yet highlight opportunities for further improvement in subsequent temporal benchmarks.

\begin{table}[htbp]
\centering
\caption{Performance comparison on event-based object tracking, a challenging yet important application for SNNs. Evaluation is conducted on two benchmark datasets, FE108 and VisEvent.}
\label{tab:fe108_visevent_tracking}
\begin{tabular}{lcc|cccc}
\toprule
\multirow{2}{*}{Architecture} & 
\multirow{2}{*}{Timesteps} & 
\multirow{2}{*}{Param. (M)} &
\multicolumn{2}{c}{FE108~\cite{zhang2021object}} &\multicolumn{2}{c}{VisEvent~\cite{wang2021viseventbenchmark}} \\
\cmidrule(lr){4-7}
& & & AUC(\%) & PR(\%)& AUC(\%) & PR(\%) \\
\midrule
SD-Track(Tiny)~\cite{shan2025sdtrackbaselineeventbasedtracking} &  $4 \times 1$ & 19.61 & 56.7&89.1 & \textbf{35.4}&48.7 \\
\cellcolor{lightblue!50} +MLPixer \((\epsilon = 4)\) &\cellcolor{lightblue!50} $4 \times 1$ &\cellcolor{lightblue!50} 20.21 &\cellcolor{lightblue!50} 57.1&\cellcolor{lightblue!50}89.2 &\cellcolor{lightblue!50} 33.7&\cellcolor{lightblue!50}47.3 \\
\cellcolor{lightblue!50} +MLPixer \((\epsilon = 6)\) &\cellcolor{lightblue!50} $4 \times 1$ & \cellcolor{lightblue!50} 22.99 &\cellcolor{lightblue!50} 57.9&\cellcolor{lightblue!50}\textbf{90.1} &\cellcolor{lightblue!50} 34.5&\cellcolor{lightblue!50}\textbf{48.9} \\
\cellcolor{lightblue} +SRB \((\epsilon = 4)\) &\cellcolor{lightblue} $4 \times 1$ &\cellcolor{lightblue} 21.43 &\cellcolor{lightblue} \textbf{58.2}&\cellcolor{lightblue}88.5 &\cellcolor{lightblue} 33.8&\cellcolor{lightblue}48.0 \\
\bottomrule
\end{tabular}
\end{table}

\section{Conclusion}
This paper presents ST-ERF as a novel framework for analyzing the spatial-temporal modeling behaviors in SNNs from a new perspective. Through this analysis, an inherent limitation in current Transformer-based SNN models is identified when applied to visual long-sequence modeling tasks. To address this limitation, two channel-mixer architectures, MLPixer and SRB, are proposed. Visualization of ST-ERF demonstrates that both modules enhance the global receptive field. Extensive experiments on long-sequence modeling tasks, including object detection and semantic segmentation, show that MLPixer and SRB improve overall performance, with SRB achieving an optimal balance between accuracy and model size. Furthermore, the study investigates complex event modeling tasks to assess the applicability of MLPixer and SRB in more challenging scenarios. Overall, the proposed ST-ERF framework offers valuable insights for the design and optimization of SNN architectures across a wide range of tasks.

\section*{Acknowledgments}
This work is supported in part by the National Natural Science Foundation of China (No. 62220106008 and 62271432), in part by the Shenzhen Science and Technology Program (Shenzhen Key Laboratory, Grant No. ZDSYS20230626091302006) and in part by the Program for Guangdong Introducing Innovative,  Entrepreneurial Teams, Grant No. 2023ZT10X044, and in part by the State Key Laboratory of Brain Cognition and Brain-inspired Intelligence Technology, Grant No. SKLBI-K2025010. This work was partially supported by UESTC Kunpeng\&Ascend Center of Cultivation.

{
\small
\bibliographystyle{unsrt}
\bibliography{ref}

\begin{thebibliography}{10}

\bibitem{MAASS19971659}
Wolfgang Maass.
\newblock Networks of spiking neurons: The third generation of neural network models.
\newblock {\em Neural Networks}, 10(9):1659--1671, 1997.

\bibitem{gerstner2002spiking}
Wulfram Gerstner and Werner~M Kistler.
\newblock {\em Spiking neuron models: Single neurons, populations, plasticity}.
\newblock Cambridge university press, 2002.

\bibitem{zhang2021rectified}
Malu Zhang, Jiadong Wang, Jibin Wu, Ammar Belatreche, Burin Amornpaisannon, Zhixuan Zhang, Venkata Pavan~Kumar Miriyala, Hong Qu, Yansong Chua, Trevor~E Carlson, et~al.
\newblock Rectified linear postsynaptic potential function for backpropagation in deep spiking neural networks.
\newblock {\em IEEE transactions on neural networks and learning systems}, 33(5):1947--1958, 2021.

\bibitem{zhang2025toward}
Malu Zhang, Shuai Wang, Jibin Wu, Wenjie Wei, Dehao Zhang, Zijian Zhou, Siying Wang, Fan Zhang, and Yang Yang.
\newblock Toward energy-efficient spike-based deep reinforcement learning with temporal coding.
\newblock {\em IEEE Computational Intelligence Magazine}, 20(2):45--57, 2025.

\bibitem{1257420}
E.M. Izhikevich.
\newblock Simple model of spiking neurons.
\newblock {\em IEEE Transactions on Neural Networks}, 14(6):1569--1572, 2003.

\bibitem{yu2014brain}
Qiang Yu, Huajin Tang, Kay~Chen Tan, and Haoyong Yu.
\newblock A brain-inspired spiking neural network model with temporal encoding and learning.
\newblock {\em Neurocomputing}, 138:3--13, 2014.

\bibitem{guo2021neural}
Wenzhe Guo, Mohammed~E Fouda, Ahmed~M Eltawil, and Khaled~Nabil Salama.
\newblock Neural coding in spiking neural networks: A comparative study for robust neuromorphic systems.
\newblock {\em Frontiers in Neuroscience}, 15:638474, 2021.

\bibitem{Qiu_Zhu_Chou_Wang_Deng_Li_2024}
Xuerui Qiu, Rui-Jie Zhu, Yuhong Chou, Zhaorui Wang, Liang-Jian Deng, and Guoqi Li.
\newblock Gated attention coding for training high-performance and efficient spiking neural networks.
\newblock {\em Proceedings of the AAAI Conference on Artificial Intelligence}, 38(1):601--610, Mar. 2024.

\bibitem{wu2018spatio}
Yujie Wu, Lei Deng, Guoqi Li, Jun Zhu, and Luping Shi.
\newblock Spatio-temporal backpropagation for training high-performance spiking neural networks.
\newblock {\em Frontiers in Neuroscience}, 12:331, 2018.

\bibitem{zhang2025memory}
Dehao Zhang, Shuai Wang, Yichen Xiao, Wenjie Wei, Yimeng Shan, Malu Zhang, and Yang Yang.
\newblock Memory-free and parallel computation for quantized spiking neural networks.
\newblock In {\em ICASSP 2025-2025 IEEE International Conference on Acoustics, Speech and Signal Processing (ICASSP)}, pages 1--5. IEEE, 2025.

\bibitem{davies2018loihi}
Mike Davies, Narayan Srinivasa, Tsung-Han Lin, Gautham Chinya, Yongqiang Cao, Sri~Harsha Choday, Georgios Dimou, Prasad Joshi, Nabil Imam, Shweta Jain, et~al.
\newblock Loihi: A neuromorphic manycore processor with on-chip learning.
\newblock {\em Ieee Micro}, 38(1):82--99, 2018.

\bibitem{pei2019towards}
Jing Pei, Lei Deng, Sen Song, Mingguo Zhao, Youhui Zhang, Shuang Wu, Guanrui Wang, Zhe Zou, Zhenzhi Wu, Wei He, et~al.
\newblock Towards artificial general intelligence with hybrid tianjic chip architecture.
\newblock {\em Nature}, 572(7767):106--111, 2019.

\bibitem{ma2024darwin3}
De~Ma, Xiaofei Jin, Shichun Sun, Yitao Li, Xundong Wu, Youneng Hu, Fangchao Yang, Huajin Tang, Xiaolei Zhu, Peng Lin, et~al.
\newblock Darwin3: a large-scale neuromorphic chip with a novel isa and on-chip learning.
\newblock {\em National Science Review}, 11(5):nwae102, 2024.

\bibitem{10378556}
Wenjie Wei, Malu Zhang, Hong Qu, Ammar Belatreche, Jian Zhang, and Hong Chen.
\newblock Temporal-coded spiking neural networks with dynamic firing threshold: Learning with event-driven backpropagation.
\newblock In {\em 2023 IEEE/CVF International Conference on Computer Vision (ICCV)}, pages 10518--10528, 2023.

\bibitem{lei2024spike2former}
Zhenxin Lei, Man Yao, Jiakui Hu, Xinhao Luo, Yanye Lu, Bo~Xu, and Guoqi Li.
\newblock Spike2former: Efficient spiking transformer for high-performance image segmentation.
\newblock {\em arXiv preprint arXiv:2412.14587}, 2024.

\bibitem{10.1007/978-3-031-73411-3_15}
Xinhao Luo, Man Yao, Yuhong Chou, Bo~Xu, and Guoqi Li.
\newblock Integer-valued training and spike-driven inference spiking neural network for high-performance and energy-efficient object detection.
\newblock In Ale{\v{s}} Leonardis, Elisa Ricci, Stefan Roth, Olga Russakovsky, Torsten Sattler, and G{\"u}l Varol, editors, {\em Computer Vision -- ECCV 2024}, pages 253--272, Cham, 2025. Springer Nature Switzerland.

\bibitem{Qiu_Yao_Zhang_Chou_Qiao_Zhou_Xu_Li_2025}
Xuerui Qiu, Man Yao, Jieyuan Zhang, Yuhong Chou, Ning Qiao, Shibo Zhou, Bo~Xu, and Guoqi Li.
\newblock Efficient 3d recognition with event-driven spike sparse convolution.
\newblock {\em Proceedings of the AAAI Conference on Artificial Intelligence}, 39(19):20086--20094, Apr. 2025.

\bibitem{shan2025sdtrackbaselineeventbasedtracking}
Yimeng Shan, Zhenbang Ren, Haodi Wu, Wenjie Wei, Rui-Jie Zhu, Shuai Wang, Dehao Zhang, Yichen Xiao, Jieyuan Zhang, Kexin Shi, Jingzhinan Wang, Jason~K. Eshraghian, Haicheng Qu, Jiqing Zhang, Malu Zhang, and Yang Yang.
\newblock Sdtrack: A baseline for event-based tracking via spiking neural networks, 2025.

\bibitem{zhu2023spikegpt}
Rui-Jie Zhu, Qihang Zhao, Guoqi Li, and Jason~K. Eshraghian.
\newblock Spikegpt: Generative pre-trained language model with spiking neural networks.
\newblock {\em arXiv preprint arXiv:2302.13939}, 2023.

\bibitem{xing2025spikellm}
Xingrun Xing, Boyan Gao, Zheng Liu, David~A. Clifton, Shitao Xiao, Wanpeng Zhang, Li~Du, Zheng Zhang, Guoqi Li, and Jiajun Zhang.
\newblock Spike{LLM}: Scaling up spiking neural network to large language models via saliency-based spiking.
\newblock In {\em The Thirteenth International Conference on Learning Representations}, 2025.

\bibitem{wang2025training}
Jingya Wang, Xin Deng, Wenjie Wei, Dehao Zhang, Shuai Wang, Qian Sun, Jieyuan Zhang, Hanwen Liu, Ning Xie, and Malu Zhang.
\newblock Training-free ann-to-snn conversion for high-performance spiking transformer.
\newblock {\em arXiv preprint arXiv:2508.07710}, 2025.

\bibitem{zhang2024spikebased}
Dehao Zhang, Shuai Wang, Ammar Belatreche, Wenjie Wei, Yichen Xiao, Haorui Zheng, Zijian Zhou, Malu Zhang, and Yang Yang.
\newblock Spike-based neuromorphic model for sound source localization.
\newblock In {\em The Thirty-eighth Annual Conference on Neural Information Processing Systems}, 2024.

\bibitem{wang2024global}
Shuai Wang, Dehao Zhang, Kexin Shi, Yuchen Wang, Wenjie Wei, Jibin Wu, and Malu Zhang.
\newblock Global-local convolution with spiking neural networks for energy-efficient keyword spotting.
\newblock {\em arXiv preprint arXiv:2406.13179}, 2024.

\bibitem{wang2025ternary}
Shuai Wang, Dehao Zhang, Ammar Belatreche, Yichen Xiao, Hongyu Qing, Wenjie Wei, Malu Zhang, and Yang Yang.
\newblock Ternary spike-based neuromorphic signal processing system.
\newblock {\em Neural Networks}, 187:107333, 2025.

\bibitem{yu2025mambaout}
Weihao Yu and Xinchao Wang.
\newblock Mambaout: Do we really need mamba for vision?
\newblock In {\em Proceedings of the IEEE/CVF Conference on Computer Vision and Pattern Recognition}, 2025.

\bibitem{7298965}
Jonathan Long, Evan Shelhamer, and Trevor Darrell.
\newblock Fully convolutional networks for semantic segmentation.
\newblock In {\em 2015 IEEE Conference on Computer Vision and Pattern Recognition (CVPR)}, pages 3431--3440, 2015.

\bibitem{vaswani2017transformer}
Ashish Vaswani, Noam Shazeer, Niki Parmar, Jakob Uszkoreit, Llion Jones, Aidan~N Gomez, {\L}ukasz Kaiser, and Illia Polosukhin.
\newblock Attention is all you need.
\newblock {\em NeurIPS}, 30, 2017.

\bibitem{dosovitskiy2020vit}
Alexey Dosovitskiy, Lucas Beyer, Alexander Kolesnikov, Dirk Weissenborn, Xiaohua Zhai, Thomas Unterthiner, Mostafa Dehghani, Matthias Minderer, Georg Heigold, Sylvain Gelly, et~al.
\newblock An image is worth 16x16 words: Transformers for image recognition at scale.
\newblock {\em arXiv preprint arXiv:2010.11929}, 2020.

\bibitem{SETR}
Sixiao Zheng, Jiachen Lu, Hengshuang Zhao, Xiatian Zhu, Zekun Luo, Yabiao Wang, Yanwei Fu, Jianfeng Feng, Tao Xiang, Philip~H.S. Torr, and Li~Zhang.
\newblock Rethinking semantic segmentation from a sequence-to-sequence perspective with transformers.
\newblock In {\em CVPR}, 2021.

\bibitem{10.1007/978-3-030-58452-8_13}
Nicolas Carion, Francisco Massa, Gabriel Synnaeve, Nicolas Usunier, Alexander Kirillov, and Sergey Zagoruyko.
\newblock End-to-end object detection with transformers.
\newblock In Andrea Vedaldi, Horst Bischof, Thomas Brox, and Jan-Michael Frahm, editors, {\em Computer Vision -- ECCV 2020}, pages 213--229, Cham, 2020. Springer International Publishing.

\bibitem{zhou2023spikformer}
Zhaokun Zhou, Yuesheng Zhu, Chao He, Yaowei Wang, Shuicheng YAN, Yonghong Tian, and Li~Yuan.
\newblock Spikformer: When spiking neural network meets transformer.
\newblock In {\em The Eleventh International Conference on Learning Representations}, 2023.

\bibitem{zhou2023spikingformerspikedrivenresiduallearning}
Chenlin Zhou, Liutao Yu, Zhaokun Zhou, Zhengyu Ma, Han Zhang, Huihui Zhou, and Yonghong Tian.
\newblock Spikingformer: Spike-driven residual learning for transformer-based spiking neural network, 2023.

\bibitem{yao2025scaling}
Man Yao, Xuerui Qiu, Tianxiang Hu, Jiakui Hu, Yuhong Chou, Keyu Tian, Jianxing Liao, Luziwei Leng, Bo~Xu, and Guoqi Li.
\newblock Scaling spike-driven transformer with efficient spike firing approximation training.
\newblock {\em IEEE Transactions on Pattern Analysis \& Machine Intelligence}, (01):1--18, 2025.

\bibitem{wang2025spiking}
Shuai Wang, Malu Zhang, Dehao Zhang, Ammar Belatreche, Yichen Xiao, Yu~Liang, Yimeng Shan, Qian Sun, Enqi Zhang, and Yang Yang.
\newblock Spiking vision transformer with saccadic attention.
\newblock In {\em The Thirteenth International Conference on Learning Representations}, 2025.

\bibitem{hubel1962receptive}
David~H Hubel and Torsten~N Wiesel.
\newblock Receptive fields, binocular interaction and functional architecture in the cat's visual cortex.
\newblock {\em The Journal of physiology}, 160(1):106, 1962.

\bibitem{le2017receptive}
Hung Le and Ali Borji.
\newblock What are the receptive, effective receptive, and projective fields of neurons in convolutional neural networks?
\newblock {\em arXiv preprint arXiv:1705.07049}, 2017.

\bibitem{NIPS2016_c8067ad1}
Wenjie Luo, Yujia Li, Raquel Urtasun, and Richard Zemel.
\newblock Understanding the effective receptive field in deep convolutional neural networks.
\newblock In D.~Lee, M.~Sugiyama, U.~Luxburg, I.~Guyon, and R.~Garnett, editors, {\em Advances in Neural Information Processing Systems}, volume~29. Curran Associates, Inc., 2016.

\bibitem{10.5555/3666122.3669371}
Zihao Wang and Lei Wu.
\newblock Theoretical analysis of inductive biases in deep convolutional networks.
\newblock In {\em Proceedings of the 37th International Conference on Neural Information Processing Systems}, NIPS '23, Red Hook, NY, USA, 2023. Curran Associates Inc.

\bibitem{MA2024102582}
Yuandong Ma, Meng Yu, Hezheng Lin, Chun Liu, Mengjie Hu, and Qing Song.
\newblock Efficient analysis of deep neural networks for vision via biologically-inspired receptive field angles: An in-depth survey.
\newblock {\em Information Fusion}, 112:102582, 2024.

\bibitem{Hayase2023UnderstandingMA}
Tomohiro Hayase and Ryo Karakida.
\newblock Understanding mlp-mixer as a wide and sparse mlp.
\newblock In {\em International Conference on Machine Learning}, 2023.

\bibitem{wei2017learning}
Zhen Wei, Yao Sun, Jinqiao Wang, Hanjiang Lai, and Si~Liu.
\newblock Learning adaptive receptive fields for deep image parsing network.
\newblock In {\em Proceedings of the IEEE conference on computer vision and pattern recognition}, pages 2434--2442, 2017.

\bibitem{gao2022rf}
Shanghua Gao, Zhong-Yu Li, Qi~Han, Ming-Ming Cheng, and Liang Wang.
\newblock Rf-next: Efficient receptive field search for convolutional neural networks.
\newblock {\em IEEE transactions on pattern analysis and machine intelligence}, 45(3):2984--3002, 2022.

\bibitem{dong2023autorf}
Peijie Dong, Xin Niu, Zimian Wei, Hengyue Pan, Dongsheng Li, and Zhen Huang.
\newblock Autorf: Auto learning receptive fields with spatial pooling.
\newblock In {\em International Conference on Multimedia Modeling}, pages 683--694. Springer, 2023.

\bibitem{mobilenetv1}
Andrew Howard, Menglong Zhu, Bo~Chen, Dmitry Kalenichenko, Weijun Wang, Tobias Weyand, Marco Andreetto, and Hartwig Adam.
\newblock Mobilenets: Efficient convolutional neural networks for mobile vision applications.
\newblock 04 2017.

\bibitem{sandler2018mobilenetv2}
Mark Sandler, Andrew Howard, Menglong Zhu, Andrey Zhmoginov, and Liang-Chieh Chen.
\newblock Mobilenetv2: Inverted residuals and linear bottlenecks.
\newblock In {\em Proceedings of the IEEE conference on computer vision and pattern recognition}, pages 4510--4520, 2018.

\bibitem{howard2019searching}
Andrew Howard, Mark Sandler, Grace Chu, Liang-Chieh Chen, Bo~Chen, Mingxing Tan, Weijun Wang, Yukun Zhu, Ruoming Pang, Vijay Vasudevan, et~al.
\newblock Searching for mobilenetv3.
\newblock In {\em Proceedings of the IEEE/CVF international conference on computer vision}, pages 1314--1324, 2019.

\bibitem{tolstikhin2021mlp}
Ilya~O Tolstikhin, Neil Houlsby, Alexander Kolesnikov, Lucas Beyer, Xiaohua Zhai, Thomas Unterthiner, Jessica Yung, Andreas Steiner, Daniel Keysers, Jakob Uszkoreit, et~al.
\newblock Mlp-mixer: An all-mlp architecture for vision.
\newblock {\em Advances in neural information processing systems}, 34:24261--24272, 2021.

\bibitem{ekambaram2023tsmixer}
Vijay Ekambaram, Arindam Jati, Nam Nguyen, Phanwadee Sinthong, and Jayant Kalagnanam.
\newblock Tsmixer: Lightweight mlp-mixer model for multivariate time series forecasting.
\newblock In {\em Proceedings of the 29th ACM SIGKDD conference on knowledge discovery and data mining}, pages 459--469, 2023.

\bibitem{ranftl2021vision}
Ren{\'e} Ranftl, Alexey Bochkovskiy, and Vladlen Koltun.
\newblock Vision transformers for dense prediction.
\newblock In {\em Proceedings of the IEEE/CVF international conference on computer vision}, pages 12179--12188, 2021.

\bibitem{10613466}
Xiangtai Li, Henghui Ding, Haobo Yuan, Wenwei Zhang, Jiangmiao Pang, Guangliang Cheng, Kai Chen, Ziwei Liu, and Chen~Change Loy.
\newblock Transformer-based visual segmentation: A survey.
\newblock {\em IEEE Transactions on Pattern Analysis and Machine Intelligence}, 46(12):10138--10163, 2024.

\bibitem{yao2024spikedriven}
Man Yao, JiaKui Hu, Tianxiang Hu, Yifan Xu, Zhaokun Zhou, Yonghong Tian, Bo~XU, and Guoqi Li.
\newblock Spike-driven transformer v2: Meta spiking neural network architecture inspiring the design of next-generation neuromorphic chips.
\newblock In {\em The Twelfth International Conference on Learning Representations}, 2024.

\bibitem{qiu2025quantized}
Xuerui Qiu, Malu Zhang, Jieyuan Zhang, Wenjie Wei, Honglin Cao, Junsheng Guo, Rui-Jie Zhu, Yimeng Shan, Yang Yang, and Haizhou Li.
\newblock Quantized spike-driven transformer.
\newblock {\em arXiv preprint arXiv:2501.13492}, 2025.

\bibitem{ding2022scaling}
Xiaohan Ding, Xiangyu Zhang, Jungong Han, and Guiguang Ding.
\newblock Scaling up your kernels to 31x31: Revisiting large kernel design in cnns.
\newblock In {\em Proceedings of the IEEE/CVF conference on computer vision and pattern recognition}, pages 11963--11975, 2022.

\bibitem{wang2025uniconvnet}
Yuhao Wang and Wei Xi.
\newblock Uniconvnet: Expanding effective receptive field while maintaining asymptotically gaussian distribution for convnets of any scale.
\newblock In {\em Proceedings of the IEEE/CVF International Conference on Computer Vision}, pages 20922--20933, 2025.

\bibitem{yao2023spikedriven}
Man Yao, JiaKui Hu, Zhaokun Zhou, Li~Yuan, Yonghong Tian, Bo~XU, and Guoqi Li.
\newblock Spike-driven transformer.
\newblock In {\em Thirty-seventh Conference on Neural Information Processing Systems}, 2023.

\bibitem{ding2021repvgg}
Xiaohan Ding, Xiangyu Zhang, Ningning Ma, Jungong Han, Guiguang Ding, and Jian Sun.
\newblock Repvgg: Making vgg-style convnets great again.
\newblock In {\em Proceedings of the IEEE/CVF conference on computer vision and pattern recognition}, pages 13733--13742, 2021.

\bibitem{NEURIPS2024_baa2da9a}
Yue Liu, Yunjie Tian, Yuzhong Zhao, Hongtian Yu, Lingxi Xie, Yaowei Wang, Qixiang Ye, Jianbin Jiao, and Yunfan Liu.
\newblock Vmamba: Visual state space model.
\newblock In A.~Globerson, L.~Mackey, D.~Belgrave, A.~Fan, U.~Paquet, J.~Tomczak, and C.~Zhang, editors, {\em Advances in Neural Information Processing Systems}, volume~37, pages 103031--103063. Curran Associates, Inc., 2024.

\bibitem{Qi2025}
Xiaoxia Qi, Md~Gapar~Md Johar, Ali Khatibi, and Jacquline Tham.
\newblock Exploiting gaussian based effective receptive fields for object detection.
\newblock {\em Scientific Reports}, 15(1):25008, July 2025.

\bibitem{li2023middle}
Chunyan Li, Zhiyong Li, Jianhong Sun, and Rui Li.
\newblock Middle-shallow feature aggregation in multimodality for face anti-spoofing.
\newblock {\em Scientific Reports}, 13(1):9870, 2023.

\bibitem{yu2024metaformer}
Weihao Yu, Chenyang Si, Pan Zhou, Mi~Luo, Yichen Zhou, Jiashi Feng, Shuicheng Yan, and Xinchao Wang.
\newblock Metaformer baselines for vision.
\newblock {\em IEEE Transactions on Pattern Analysis and Machine Intelligence}, 46(2):896--912, 2024.

\bibitem{lin2014ms-coco}
Tsung-Yi Lin, Michael Maire, Serge Belongie, James Hays, Pietro Perona, Deva Ramanan, Piotr Doll{\'a}r, and C~Lawrence Zitnick.
\newblock Microsoft coco: Common objects in context.
\newblock In {\em Computer Vision--ECCV 2014: 13th European Conference, Zurich, Switzerland, September 6-12, 2014, Proceedings, Part V 13}, pages 740--755. Springer, 2014.

\bibitem{chen1906mmdetection}
Kai Chen, Jiaqi Wang, Jiangmiao Pang, Yuhang Cao, Yu~Xiong, Xiaoxiao Li, Shuyang Sun, Wansen Feng, Ziwei Liu, Jiarui Xu, et~al.
\newblock Mmdetection: Open mmlab detection toolbox and benchmark. arxiv 2019.
\newblock {\em arXiv preprint arXiv:1906.07155}, 5, 1906.

\bibitem{he2017mask}
Kaiming He, Georgia Gkioxari, Piotr Doll{\'a}r, and Ross Girshick.
\newblock Mask r-cnn.
\newblock In {\em Proceedings of the IEEE international conference on computer vision}, pages 2961--2969, 2017.

\bibitem{deng2009imagenet}
Jia Deng, Wei Dong, Richard Socher, Li-Jia Li, Kai Li, and Li~Fei-Fei.
\newblock Imagenet: A large-scale hierarchical image database.
\newblock In {\em CVPR}, pages 248--255. Ieee, 2009.

\bibitem{glorot2010understanding}
Xavier Glorot and Yoshua Bengio.
\newblock Understanding the difficulty of training deep feedforward neural networks.
\newblock In {\em Proceedings of the thirteenth international conference on artificial intelligence and statistics}, pages 249--256. JMLR Workshop and Conference Proceedings, 2010.

\bibitem{zhou2019ade20k}
Bolei Zhou, Hang Zhao, Xavier Puig, Tete Xiao, Sanja Fidler, Adela Barriuso, and Antonio Torralba.
\newblock Semantic understanding of scenes through the ade20k dataset.
\newblock {\em IJCV}, 127:302--321, 2019.

\bibitem{contributors2020mmsegmentation}
MMSegmentation Contributors.
\newblock Mmsegmentation: Openmmlab semantic segmentation toolbox and benchmark, 2020.

\bibitem{kirillov2019semetic-fpn}
Alexander Kirillov, Ross Girshick, Kaiming He, and Piotr Doll{\'a}r.
\newblock Panoptic feature pyramid networks.
\newblock In {\em CVPR}, pages 6399--6408, 2019.

\bibitem{zhang2021object}
Jiqing Zhang, Xin Yang, Yingkai Fu, Xiaopeng Wei, Baocai Yin, and Bo~Dong.
\newblock Object tracking by jointly exploiting frame and event domain.
\newblock In {\em Proceedings of the IEEE/CVF International Conference on Computer Vision}, pages 13043--13052, 2021.

\bibitem{wang2021viseventbenchmark}
Xiao Wang, Jianing Li, Lin Zhu, Zhipeng Zhang, Zhe Chen, Xin Li, Yaowei Wang, Yonghong Tian, and Feng Wu.
\newblock Visevent: Reliable object tracking via collaboration of frame and event flows.
\newblock {\em arXiv:2108.05015}, 2021.

\bibitem{xiao2022online}
Mingqing Xiao, Qingyan Meng, Zongpeng Zhang, Di~He, and Zhouchen Lin.
\newblock Online training through time for spiking neural networks.
\newblock {\em Advances in Neural Information Processing Systems}, 35:20717--20730, 2022.

\bibitem{mmdetection}
Kai Chen, Jiaqi Wang, Jiangmiao Pang, Yuhang Cao, Yu~Xiong, Xiaoxiao Li, Shuyang Sun, Wansen Feng, Ziwei Liu, Jiarui Xu, Zheng Zhang, Dazhi Cheng, Chenchen Zhu, Tianheng Cheng, Qijie Zhao, Buyu Li, Xin Lu, Rui Zhu, Yue Wu, Jifeng Dai, Jingdong Wang, Jianping Shi, Wanli Ouyang, Chen~Change Loy, and Dahua Lin.
\newblock {MMDetection}: Open mmlab detection toolbox and benchmark.
\newblock {\em arXiv preprint arXiv:1906.07155}, 2019.

\bibitem{mmseg2020}
MMSegmentation Contributors.
\newblock {MMSegmentation}: Openmmlab semantic segmentation toolbox and benchmark.
\newblock \url{https://github.com/open-mmlab/mmsegmentation}, 2020.

\end{thebibliography}
}

\newpage
\section*{NeurIPS Paper Checklist}

\begin{enumerate}

\item {\bf Claims}
    \item[] Question: Do the main claims made in the abstract and introduction accurately reflect the paper's contributions and scope?
    \item[] Answer: \answerYes{} 
    \item[] Justification: Our abstract and introduction clearly describe our contribution, the algorithm, and the experimental results.

\item {\bf Limitations}
    \item[] Question: Does the paper discuss the limitations of the work performed by the authors?
    \item[] Answer: \answerYes{} 
    \item[] Justification: This paper discusses the limitations of the work in Appendix~\ref{Limitations}.

\item {\bf Theory assumptions and proofs}
    \item[] Question: For each theoretical result, does the paper provide the full set of assumptions and a complete (and correct) proof?
    \item[] Answer: \answerYes{} 
    \item[] Justification: We mentioned our proposition and properties of ST-ERF and provided a whole set of proofs in Appendix \ref{proof_property_1}.

    \item {\bf Experimental result reproducibility}
    \item[] Question: Does the paper fully disclose all the information needed to reproduce the main experimental results of the paper to the extent that it affects the main claims and/or conclusions of the paper (regardless of whether the code and data are provided or not)?
    \item[] Answer: \answerYes{} 
    \item[] Justification: We provide a detailed description of our model architecture and present all the training details, including dataset processing methods and hyperparameter settings in Appendix~\ref{detail_det_seg}.

\item {\bf Open access to data and code}
    \item[] Question: Does the paper provide open access to the data and code, with sufficient instructions to faithfully reproduce the main experimental results, as described in supplemental material?
    \item[] Answer: \answerYes{} 
    \item[] Justification: We mentioned our data in Appendix~\ref{detail_det_seg}. The code is compressed in the supplemental material.

\item {\bf Experimental setting/details}
    \item[] Question: Does the paper specify all the training and test details (e.g., data splits, hyperparameters, how they were chosen, type of optimizer, etc.) necessary to understand the results?
    \item[] Answer: \answerYes{} 
    \item[] Justification: We provide the full details in Appendix~\ref{detail_det_seg}.

\item {\bf Experiment statistical significance}
    \item[] Question: Does the paper report error bars suitably and correctly defined or other appropriate information about the statistical significance of the experiments?
    \item[] Answer: \answerYes{} 
    \item[] Justification: Although we did provided $mean \pm std$ range to show several experiments' numerical range, we need to clarify that the numerical experiments are focused on the validation of ST-ERF properties, so we provided several independent trials with different input samples and different network initialization. The mean value of performance (e.g. the fitted curve) is solid enough to clarify our theories. 

\item {\bf Experiments compute resources}
    \item[] Question: For each experiment, does the paper provide sufficient information on the computer resources (type of compute workers, memory, time of execution) needed to reproduce the experiments?
    \item[] Answer: \answerYes{} 
    \item[] Justification: This paper provides sufficient information on the computer resources.
    
\item {\bf Code of ethics}
    \item[] Question: Does the research conducted in the paper conform, in every respect, with the NeurIPS Code of Ethics \url{https://neurips.cc/public/EthicsGuidelines}?
    \item[] Answer: \answerYes{} 
    \item[] Justification: This paper strictly adheres to the NeurIPS Code of Ethics.

\item {\bf Broader impacts}
    \item[] Question: Does the paper discuss both potential positive societal impacts and negative societal impacts of the work performed?
    \item[] Answer: \answerNA{} 
    \item[] Justification: This work is a foundational research and not tied to particular societal applications.
    
\item {\bf Safeguards}
    \item[] Question: Does the paper describe safeguards that have been put in place for responsible release of data or models that have a high risk for misuse (e.g., pretrained language models, image generators, or scraped datasets)?
    \item[] Answer: \answerNA{} 
    \item[] Justification: This paper poses no such risks.

\item {\bf Licenses for existing assets}
    \item[] Question: Are the creators or original owners of assets (e.g., code, data, models), used in the paper, properly credited and are the license and terms of use explicitly mentioned and properly respected?
    \item[] Answer: \answerYes{} 
    \item[] Justification: Yes, the paper properly credits the creators or original owners of assets (e.g., code, data, models) and explicitly mentions and respects the relevant licenses and terms of use.

\item {\bf New assets}
    \item[] Question: Are new assets introduced in the paper well documented and is the documentation provided alongside the assets?
    \item[] Answer: \answerNA{} 
    \item[] Justification: No new assets are introduced in this article.

\item {\bf Crowdsourcing and research with human subjects}
    \item[] Question: For crowdsourcing experiments and research with human subjects, does the paper include the full text of instructions given to participants and screenshots, if applicable, as well as details about compensation (if any)? 
    \item[] Answer: \answerNA{} 
    \item[] Justification: This paper does not involve crowdsourcing nor research with human subjects.

\item {\bf Institutional review board (IRB) approvals or equivalent for research with human subjects}
    \item[] Question: Does the paper describe potential risks incurred by study participants, whether such risks were disclosed to the subjects, and whether Institutional Review Board (IRB) approvals (or an equivalent approval/review based on the requirements of your country or institution) were obtained?
    \item[] Answer: \answerNA{} 
    \item[] Justification: The paper does not involve crowdsourcing nor research with human subjects.

\item {\bf Declaration of LLM usage}
    \item[] Question: Does the paper describe the usage of LLMs if it is an important, original, or non-standard component of the core methods in this research? Note that if the LLM is used only for writing, editing, or formatting purposes and does not impact the core methodology, scientific rigorousness, or originality of the research, declaration is not required.
    \item[] Answer: \answerNA{} 
    \item[] Justification: The proposed method in this research does not involve LLMs as any important, original, or non-standard components.

\end{enumerate}

\newpage
\appendix
\section*{Appendix}

\section{Proof of~\ref{LDC}}
\label{proof_property_1}
\begin{proof}
    Leaky Integrate-and-Fire (LIF) model \cite{1257420, gerstner2002spiking} can be described by the following equations:
    \begin{align}
    \label{eq:mem} 
    \mathbf{v}^{\ell}[t]&=\mathbf{h}^{\ell}[t-1]+f({\mathbf{w}^{\ell}},\mathbf{x}^{\ell-1}[t-1]), &\text{(Charging function)}, \\
    \label{eq:lif}
    \mathbf{s}^{\ell}[t]&=\mathbf{\Theta}(\mathbf{v}^{\ell}[t]-\vartheta), &\text{(Firing function)},\\
    \label{eq:reset} 
    \mathbf{h}^{\ell}[t]&= \begin{cases}
        \beta\mathbf{v}^{\ell}[t]- \vartheta \mathbf{s}^{\ell}[t], &\text{soft reset} \\
        \mathbf{v}^{\ell}[t](1- \mathbf{s}^{\ell}[t]), &\text{hard reset}
    \end{cases}
        &\text{(Leak-and-reset function),}
    \end{align}
    where $\beta$ is the decay constant, $t$ is the time step, $\mathbf{w}^{\ell}$ is the weight matrix of layer $\ell$, $f(\cdot) $ is the operation that stands for convolution (Conv) or fully connected (FC), $\bf x$ is the input, and  $\mathbf{\Theta(\cdot)}$ denotes the Heaviside step function. When the membrane potential $\mathbf{v}$ exceeds the firing threshold $\vartheta$, the LIF neuron will trigger a spike $\mathbf{s}$; otherwise, it remains inactive. After spike emission, the neuron invokes the reset mechanism, where the soft reset function is employed. $\bf h$ is the membrane potential following the reset function. 
    
    For the back-propagation of this neuron, we introduce the training process of SNN gradient descent and the parameter update method of spatio-temporal back-propagation (STBP) \cite{wu2018spatio,xiao2022online}. The accumulated gradients of loss $\mathcal{L}$ with respect to weights $\mathbf{w}$ at layer $\ell$ can be calculated as:
    {\small
    \begin{equation}
        \begin{aligned}
\frac{\partial \mathcal{L}}{\partial \mathbf{w}^{\ell}} = \sum_{t=1}^T \frac{\partial \mathcal{L}}{\partial \mathbf{s}^{\ell+1}[t]}\frac{\partial \mathbf{s}^{\ell+1}[t]}{\partial \mathbf{v}^{\ell+1}[t]}(\frac{\partial \mathbf{v}^{\ell+1}[t]}{\partial \mathbf{w}^\ell} + \sum_{\tau < t}\prod_{i=t-1}^{\tau}\left({\frac{\partial \mathbf{v}^{\ell+1}[i+1]}{\partial \mathbf{v}^{\ell+1}[i]}}+{\frac{\partial \mathbf{v}^{\ell+1}[i+1]}{\partial \mathbf{s}^{\ell+1}[i]}}{\frac{\partial \mathbf{s}^{\ell+1}[i]}{\partial \mathbf{v}^{\ell+1}[i]}}\right)\frac{\partial \mathbf{v}^{\ell+1}[\tau]}{\partial \mathbf{w}^\ell}),
\end{aligned}
    \end{equation}
    }
    
    where $\mathbf{s}^{\ell}[t]$ and $\mathbf{v}^{\ell}[t]$ represent the output spikes and membrane potential of the neuron in layer $\ell$, at time $t$. Moreover, notice that $ \frac{\partial \mathbf{s}^{\ell}[t]}{\partial \mathbf{v}^{\ell}[t]}$ is non-differentiable. To overcome this problem, Wu et al. \cite{wu2018spatio} propose the surrogate function to make only the neurons whose membrane potentials close to the firing threshold receive nonzero gradients during back-propagation.  
    
    In this paper, we use the rectangle function, which has been shown to be effective in gradient descent and may be calculated by:
    \begin{equation}
    \label{eq3}
        \frac{\partial \mathbf{s}^{\ell}[t]}{\partial \mathbf{v}^{\ell}[t]}=\frac{1}{a} \operatorname{sign}\left(\left|\mathbf{v}^{\ell}[t]-\vartheta \right|<\frac{a}{2}\right),
    \end{equation}
    where $a$ is a defined coefficient for controlling the width of the gradient window.

    To compute $\sum_{t=1}^{T} \frac{\partial \mathbf{s}^{\ell}_{(0,0)}[t]}{\partial \mathbf{s}^{\ell-1}_{(i,j)}[t]}$, we follow the chain rule with an arbitrary loss $\mathcal{L}$. Consider $\sum_{t=1}^{T} \frac{\partial \mathcal{L}}{\partial \mathbf{s}^{\ell-1}_{(i,j)}[t]}$:

    \begin{equation}
        \begin{aligned}
            \sum_{t=1}^{T} \frac{\partial \mathcal{L}}{\partial \mathbf{s}^{\ell-1}_{(i,j)}[t]} &= \sum_{t=1}^{T} \sum_{\hat{i}, \hat{j}} \frac{\partial \mathcal{L}}{\partial \mathbf{s}^{\ell}_{(\hat{i},\hat{j})}[t]} \frac{\partial \mathbf{s}^{\ell}_{(\hat{i},\hat{j})}[t]}{\partial \mathbf{s}^{\ell-1}_{(i,j)}[t]}\\
            &= \sum_{\hat{i}} \sum_{\hat{j}} \sum_{t=1}^{T} \frac{\partial \mathcal{L}}{\partial \mathbf{s}^{\ell}_{(\hat{i},\hat{j})}[t]} \frac{\partial \mathbf{s}^{\ell}_{(\hat{i},\hat{j})}[t]}{\partial \mathbf{s}^{\ell-1}_{(i,j)}[t]} \\
            &= \sum_{\hat{i} \neq 0} \sum_{\hat{j} \neq 0} \sum_{t=1}^{T} \frac{\partial \mathcal{L}}{\partial \mathbf{s}^{\ell}_{(\hat{i},\hat{j})}[t]} \frac{\partial \mathbf{s}^{\ell}_{(\hat{i},\hat{j})}[t]}{\partial \mathbf{s}^{\ell-1}_{(i,j)}[t]} + \sum_{t=1}^{T} \frac{\partial \mathcal{L}}{\partial \mathbf{s}^{\ell}_{(0,0)}[t]} \frac{\partial \mathbf{s}^{\ell}_{(0,0)}[t]}{\partial \mathbf{s}^{\ell-1}_{(i,j)}[t]}.
        \end{aligned}
    \end{equation}

    When the following conditions are met:
    \begin{equation}
        \label{cond}
        \forall t \in T, \frac{\partial \mathcal{L}}{\partial \mathbf{s}^{\ell}_{(\hat{i},\hat{j})}[t]} =  
        \begin{cases}
            1 & \hat{i}=0,\hat{j}=0, \\
            0 & otherwise
        \end{cases}.
    \end{equation}

    We can get:
    \begin{equation}
        \begin{aligned}
        \sum_{t=1}^{T} \frac{\partial \mathcal{L}}{\partial \mathbf{s}^{\ell-1}_{(i,j)}[t]} = \sum_{t=1}^{T} \frac{\partial \mathbf{s}^{\ell}_{(0,0)}[t]}{\partial \mathbf{s}^{\ell-1}_{(i,j)}[t]}, \quad 
        \frac{1}{T} \sum_{t=1}^{T} \frac{\partial \mathcal{L}}{\partial \mathbf{s}^{\ell-1}_{(i,j)}[t]} = \frac{1}{T} \sum_{t=1}^{T} \frac{\partial \mathbf{s}^{\ell}_{(0,0)}[t]}{\partial \mathbf{s}^{\ell-1}_{(i,j)}[t]}.
        \end{aligned}
    \end{equation}

    The spatial ERF at position $(i,j)$ can thus be calculated by summing the gradients of the loss with respect to all timesteps.

    For the temporal ERF, we need to compute $\sum_{i,j} \frac{\partial s_{(0,0)}^{\ell}[T]}{\partial s_{(i,j)}^{\ell-1}[T-\tau]}$. We consider $\sum_{i,j} \frac{\partial \mathcal{L}}{\partial \mathbf{s}_{(i,j)}^{\ell-1}[T-\tau]}$. By applying the chain rule:
    
    \begin{equation}
        \begin{aligned}
        \sum_{i,j} \frac{\partial \mathcal{L}}{\partial \mathbf{s}_{(i,j)}^{\ell-1}[T-\tau]} &= \sum_{i,j} \sum_{\hat{i},\hat{j}} \frac{\partial \mathcal{L}}{\partial \mathbf{s}_{(\hat{i},\hat{j})}^{\ell}[T]} \frac{\partial \mathbf{s}_{(\hat{i},\hat{j})}^{\ell}[T]}{\partial \mathbf{s}_{(i,j)}^{\ell-1}[T-\tau]} \\
        \end{aligned}
    \end{equation}    

    When the following conditions are met:
    \begin{equation}
        \forall \hat{i},\hat{j}, \frac{\partial \mathcal{L}}{\partial \mathbf{s}^{\ell}_{(\hat{i},\hat{j})}[T]} = 1 
    \end{equation}

    We can simplify:
    \begin{equation}
        \sum_{i,j} \frac{\partial \mathcal{L}}{\partial s_{(i,j)}^{\ell-1}[T-\tau]} = \sum_{\hat{i},\hat{j}}\sum_{i,j} \frac{\partial s_{(\hat{i},\hat{j})}^{\ell}[T]}{\partial s_{(i,j)}^{\ell-1}[T-\tau]}
    \end{equation}

    The temporal ERF at delay $\tau$ can thus be calculated by summing the gradients of the loss with respect to all spatial positions at timestep $T-\tau$.
\end{proof}

\section{Details in Detection and Segmentation Experiments}
\label{detail_det_seg}
\begin{table}[!h]
\caption{Configurations of different Meta-SDT Variants.}
\centering
\small
\label{table_imagenet_config}
\vspace{0pt}
\renewcommand{\arraystretch}{1.2}
\setlength{\tabcolsep}{3pt}
\begin{tabular}{c|c|ccc|ccc}
\hline
Stage & \# Tokens & \multicolumn{3}{c|}{Layer Specification}& \multicolumn{1}{c|}{Tiny} & \multicolumn{1}{c|}{Medium} & Base \\ \hline

\multirow{14}{*}{1} & \multirow{7}{*}{$\dfrac{H}{2}$ \texttimes $\dfrac{W}{2}$}   & \multicolumn{2}{c|}{\multirow{2}{*}{Downsampling}} & Conv & \multicolumn{3}{c}{7x7 stride 2}
\\ \cline{5-8} 
& & \multicolumn{2}{c|}{} & Dim & \multicolumn{1}{c|}{16}  & \multicolumn{1}{c|}{24}  & 32  
\\ \cline{3-8} 
& & \multicolumn{1}{c|}{\multirow{5}{*}{\begin{tabular}[c]{@{}c@{}}Conv-based\\      SNN block\end{tabular}}}        & \multicolumn{1}{c|}{\multirow{2}{*}{SepConv}}      & DWConv     & \multicolumn{3}{c}{7x7 stride 1}                          
\\ \cline{5-8} 
& & \multicolumn{1}{c|}{} & \multicolumn{1}{c|}{} & MLP ratio  & \multicolumn{3}{c}{2} 
\\ \cline{4-8} 
& & \multicolumn{1}{c|}{} & \multicolumn{1}{c|}{\multirow{3}{*}{Channel Mixer}} &(1)Conv+Conv       & \multicolumn{3}{c}{$\mathrm{\epsilon}=4$}                        
\\ \cline{5-8} 
& & \multicolumn{1}{c|}{} & \multicolumn{1}{c|}{} &(2)MLP+MLP & \multicolumn{3}{c}{$\mathrm{\epsilon}=4/6$}                \\ \cline{5-8}
& & \multicolumn{1}{c|}{} & \multicolumn{1}{c|}{} & (3)MLP+Conv & \multicolumn{3}{c}{$\mathrm{\epsilon}=4$} \\
\cline{2-8} 
& \multirow{7}{*}{$\dfrac{H}{4}$ \texttimes $\dfrac{W}{4}$}   & \multicolumn{2}{c|}{\multirow{2}{*}{Downsampling}} & Conv       & \multicolumn{3}{c}{3x3 stride 2}            \\ \cline{5-8} 
& & \multicolumn{2}{c|}{} & Dim        & \multicolumn{1}{c|}{32}  & \multicolumn{1}{c|}{48}  & 64
\\ \cline{3-8} 
& & \multicolumn{1}{c|}{\multirow{5}{*}{\begin{tabular}[c]{@{}c@{}}Conv-based\\      SNN block\end{tabular}}}        & \multicolumn{1}{c|}{\multirow{2}{*}{SepConv}}      & DWConv     & \multicolumn{3}{c}{7x7 stride 1}                          
\\ \cline{5-8} 
& & \multicolumn{1}{c|}{} & \multicolumn{1}{c|}{}                              & MLP ratio  & \multicolumn{3}{c}{2}                                     
\\ \cline{4-8} 
& & \multicolumn{1}{c|}{} & \multicolumn{1}{c|}{\multirow{3}{*}{Channel Mixer}} & (1)Conv+Conv       & \multicolumn{3}{c}{$\mathrm{\epsilon}=4$}                        \\ \cline{5-8} 
& & \multicolumn{1}{c|}{} & \multicolumn{1}{c|}{} & (2)MLP+MLP & \multicolumn{3}{c}{$\mathrm{\epsilon}=4/6$}    \\ \cline{5-8}   
& & \multicolumn{1}{c|}{} & \multicolumn{1}{c|}{} & (3)MLP+Conv & \multicolumn{3}{c}{$\mathrm{\epsilon}=4$} 
\\ \hline
\multirow{7}{*}{2}  & \multirow{7}{*}{$\dfrac{H}{8}$ \texttimes $\dfrac{W}{8}$}   & \multicolumn{2}{c|}{\multirow{2}{*}{Downsampling}} & Conv       & \multicolumn{3}{c}{3x3 stride 2}                          
\\ \cline{5-8} 
& & \multicolumn{2}{c|}{} & Dim & \multicolumn{1}{c|}{64} & \multicolumn{1}{c|}{96} & 128 
\\ \cline{3-8} 
& & \multicolumn{1}{c|}{\multirow{6}{*}{\begin{tabular}[c]{@{}c@{}}Conv-based\\      SNN block\end{tabular}}}        & \multicolumn{1}{c|}{\multirow{2}{*}{SepConv}}      & DWConv     & \multicolumn{3}{c}{7x7 stride 1}                          
\\ \cline{5-8} 
& & \multicolumn{1}{c|}{} & \multicolumn{1}{c|}{}                              & MLP ratio  & \multicolumn{3}{c}{2}                                     
\\ \cline{4-8} & & \multicolumn{1}{c|}{} & \multicolumn{1}{c|}{\multirow{3}{*}{Channel Mixer}} & (1)Conv+Conv       & \multicolumn{3}{c}{$\mathrm{\epsilon}=4$}          \\ \cline{5-8} 
& & \multicolumn{1}{c|}{} & \multicolumn{1}{c|}{}                              & (2)MLP+MLP & \multicolumn{3}{c}{$\mathrm{\epsilon}=4/6$}            \\ \cline{5-8} 
& & \multicolumn{1}{c|}{} & \multicolumn{1}{c|}{}       & (3)MLP+Conv & \multicolumn{3}{c}{$\mathrm{\epsilon}=4$}
\\ \cline{4-8} 
& & \multicolumn{1}{c|}{} & \multicolumn{2}{c|}{\# Blocks} & \multicolumn{3}{c}{2}                                    
\\ \hline
\multirow{5}{*}{3}  & \multirow{5}{*}{$\dfrac{H}{16}$ \texttimes $\dfrac{W}{16}$} & \multicolumn{2}{c|}{\multirow{2}{*}{Downsampling}} & Conv       & \multicolumn{3}{c}{3x3 stride 2}                         
\\ \cline{5-8} 
& & \multicolumn{2}{c|}{} & Dim        & \multicolumn{1}{c|}{128} & \multicolumn{1}{c|}{192} & 256
\\ \cline{3-8} 
& & \multicolumn{1}{c|}{\multirow{3}{*}{\begin{tabular}[c]{@{}c@{}}Transformer-based\\      SNN block\end{tabular}}} & \multicolumn{1}{c|}{SDSA}                          & RepConv    & \multicolumn{3}{c}{3x3 stride 1}                          
\\ \cline{4-8} 
& & \multicolumn{1}{c|}{} & \multicolumn{1}{c|}{Channel MLP}                   & MLP ratio  & \multicolumn{3}{c}{4}                                     
\\ \cline{4-8}  
& & \multicolumn{1}{c|}{} & \multicolumn{2}{c|}{\# Blocks} & \multicolumn{3}{c}{6}                                     
\\ \hline
\multirow{5}{*}{4}  & \multirow{5}{*}{$\dfrac{H}{16}$ \texttimes $\dfrac{W}{16}$} & \multicolumn{2}{c|}{\multirow{2}{*}{Downsampling}} & Conv       & \multicolumn{3}{c}{3x3 stride 1}                         
\\ \cline{5-8} 
& & \multicolumn{2}{c|}{}  & Dim & \multicolumn{1}{c|}{192} & \multicolumn{1}{c|}{240} & 360
\\ \cline{3-8} 
&  & \multicolumn{1}{c|}{\multirow{3}{*}{\begin{tabular}[c]{@{}c@{}}Transformer-based\\      SNN block\end{tabular}}} & \multicolumn{1}{c|}{SDSA}                          & RepConv    & \multicolumn{3}{c}{3x3 stride 1}                          
\\ \cline{4-8} 
& & \multicolumn{1}{c|}{} & \multicolumn{1}{c|}{Channel MLP}                   & MLP ratio  & \multicolumn{3}{c}{4}                                     
\\ \cline{4-8} 
& & \multicolumn{1}{c|}{} & \multicolumn{2}{c|}{\# Blocks} & \multicolumn{3}{c}{2}                                    
\\ \hline
\end{tabular}
\end{table}

\begin{table}[h]
\vspace{-0.2cm}
\centering
\small
\caption{Hyper-parameters for pre-training on ImageNet-1K}
\label{table_train_imagenet_detail}
\begin{tabular}{c|c||c|c}
\hline
Hyper-parameter     & Settings  & Hyper-parameter & Settings   \\ \hline
Model size          & T/M/B     &Timestemp    & 4       \\
Epochs              & 200       &Resolution   & 224*224 \\
Batch size          & 1568      &Optimizer    & LAMB \\
Base learning rate  & 6e-4   &Learning rate decay & Cosine \\
Warmup eopchs       & 10     &Weight decay        & 0.05\\
Random augment        & 9/0.5  &Mixup               & None \\
Cutmix              & None   &Label smoothing     & 0.1  \\ \hline
\end{tabular}
\end{table}

On ImageNet-1K pretraining, we employ three scales of Meta-SDT with three different channel mixer design ((1): Conv-Mixer; (2): MLPixer; (3): SRB) in Table~\ref{table_imagenet_config} and utilize the hyper-parameters in Table~\ref{table_train_imagenet_detail} to pre-train models in our paper for further fine-tuning on COCO 2017 and ADE20K datasets. Note that $\epsilon$ represents the channel expand ratio (CHW $\rightarrow$ $\epsilon$ CHW $\rightarrow$ CHW).

For COCO 2017 dataset, We utilize the MMDetection~\cite{mmdetection} framework to implement the existing models and our method. The object detection and instance segmentation framework strictly follows Mask R-CNN, with a training schedule of $1 \times$ (12 epochs). We use a total batch size of 4/GPU, utilize the AdamW optimizer with a learning rate of $1 \times 10^{-4}$ and a weight decay of 0.05. Images are resized and cropped into 1333 × 800 for training and testing and maintain the ratio. Random horizontal flipping and resize with a ratio of 0.5 was applied for augmentation during training. This pre-training fine-tuning method is a commonly used strategy in ANNs.

For ADE20K dataset, we utilize the MMSegmentation~\cite{mmseg2020} framework. The training configuration strictly encompasses for 160,000 iterations. The batch size is set to 4/GPU, and the AdamW optimizer is used. The learning rate and weight decay parameters are tuned to $2 \times 10^{-4}$ and 0.05, respectively. To speed up training, we warm up the model for 1.5k iterations with a linear decay schedule. All the experiments are conducted on 4 NVIDIA-A100 80GB GPUs.

\newpage
\section{Limitations}
\label{Limitations}
This work presents several avenues for future exploration, such as how neuronal dynamics parameters influence ST-ERF in more dynamic and diverse SNNs. Given that one of SNN's major successes stems from its inherent membrane potential memory update mechanism, this represents a particularly worthwhile direction for deeper investigation. We will further explore the interactions between spiking neurons' neurodynamics and the networks' temporal response in the future. Nevertheless, this work provides a viable analytical framework for understanding SNN model behavior, with practical implications for architectural design across various levels of SNNs.

\end{document}